\documentclass[11pt]{article}

\usepackage[utf8]{inputenc}
\usepackage[T1]{fontenc}
\usepackage{lmodern}
\usepackage[margin=1in]{geometry}
\usepackage{microtype}
\usepackage{amsmath,amssymb}
\usepackage{booktabs}
\usepackage{array}
\usepackage{tabularx}
\usepackage{longtable}
\usepackage{float}
\usepackage[section]{placeins}
\usepackage{graphicx}
\usepackage{xcolor}
\usepackage{tikz}
\usetikzlibrary{arrows.meta,positioning,fit,backgrounds,calc}
\usepackage{authblk}
\usepackage{enumitem}
\usepackage{caption}
\usepackage[hidelinks]{hyperref}
\usepackage{url}

\definecolor{verG}{HTML}{1B7A3D}
\definecolor{gateR}{HTML}{B23A1E}
\definecolor{boxbg}{HTML}{F4F5F7}
\definecolor{pbPale}{HTML}{EDF1FE}
\definecolor{pbMid}{HTML}{C7D2F6}
\definecolor{pbDeep}{HTML}{8B9BE8}
\definecolor{pbEdge}{HTML}{7C8CCB}
\newcommand{\ver}{\textcolor{verG}{\textbf{verified}}}
\newcommand{\gat}{\textcolor{gateR}{\textbf{gated}}}
\newcommand{\mh}{\textsf{materialHash}}
\newcommand{\hh}{\textsf{headHash}}
\newcommand{\sfh}{\textsf{storyFilesHash}}
\newcommand{\code}[1]{\texttt{\small #1}}
\newcolumntype{Y}{>{\raggedright\arraybackslash}X}

\title{
Proof-or-Stop:\\
Don't Trust the Agent, Trust the Evidence\\[0.4em]
\large Loop Engineering for Verifiable Evidence-Gated Lifecycle Control
}

\author{Jek Huang\thanks{Corresponding author: \texttt{jek.huang@prodenovo.ai}}, Jeffery Hsia, Jiayi Sun, Freddie Shi, Wei Huang, and Ian H. White}
\date{June 2026 \quad (preprint v1)}

\begin{document}
\maketitle

\begin{abstract}
\noindent
Autonomous coding agents increasingly execute multi-step software work. However, lifecycle states
such as reviewed, tested, \textsc{done}, and ready-to-merge remain claims unless a downstream
system can decide whether those claims are supported by current evidence. In this work, we
present \textbf{Proof-or-Stop Lifecycle Control}, a method in which lifecycle transitions are
admitted only when fresh, tracked-source-state-bound, mechanically verifiable evidence satisfies
the relevant gate. The method instantiates an \emph{agent-as-claim} lifecycle semantics: agent
outputs may propose lifecycle claims, but do not themselves constitute lifecycle state. Here,
``proof'' is used operationally to mean gate-admissible evidence under a stated trust model, not a
proof of semantic program correctness.

The method was instantiated in the open-source \emph{Proof-or-Stop} implementation and evaluated
through mechanism tests, a powered control-policy ablation, and operated self-application
evidence. Mechanism checks show that \textsc{done} and receipt claims do not advance on
self-report in the tested engine contract: the unattended-loop engine passed 10/10 scenarios with
zero false-\textsc{done}, and local-key receipt bundles rejected 18 tamper classes with zero false
accepts in the tested suite. In a 9{,}240-cell powered ablation, the pre-registered A4-vs-A2$'$
contrast reduced visible-pass/hidden-fail amplification from $31/1800$ injected cells under a
compute-budgeted naive loop to $2/1800$ under the gated loop (+1.6pp not-amplified, 95\% CI
$[0.8,2.5]$). The separation was concentrated in a trap-active task; the near-compute A3-vs-A4
contrast ($14/1800$ vs $2/1800$) indicates that the improvement is associated with enforcing the
review signal as a lifecycle gate, rather than merely adding a reviewer. Finally, the operated
self-application corpus (\textbf{565 stories / 1007 review findings}, 94.8\% resolved) and a
refreshed \textbf{68-row high/critical} cross-vendor exhibit show that the system produces
auditable evidence on its own development. Together, these results support Proof-or-Stop as a
model-agnostic, host-neutral control layer for deciding which autonomous-agent claims a lifecycle
may safely act on, rather than as a new model or coding agent. The evaluation is limited to one
model family, 24 ablation tasks, and a self-hosted corpus.
\end{abstract}


\section{Introduction}

Autonomous coding systems increasingly combine durable execution, tool use, review agents, and
handoff protocols. These systems have made agent work more executable, resumable, and measurable.
However, they do not by themselves decide whether lifecycle claims such as \emph{tested},
\emph{reviewed}, \emph{done}, and \emph{ready to merge} are safe for downstream automation to
act on. This distinction becomes load-bearing when an unattended agent can generate code, retry
until visible checks pass, and narrate completion in the same workflow that will be asked to
advance the work.

The resulting problem is claim admissibility. A self-report is not evidence; a log line saying
``All tests passed'' is not evidence that the tests correspond to the code about to be merged;
and a reviewer response saying ``LGTM'' is not, by itself, an artifact that a later gate can
re-check. In existing practice, a green pipeline or a successful agent handoff can therefore
coexist with a stale, incomplete, or unsupported lifecycle claim. The missing control is not
another model, but an admissibility rule for deciding when a claim may move lifecycle state.

\paragraph{Thesis.} In this work, we present \textbf{Proof-or-Stop Lifecycle Control}. The key
idea is to treat every consequential actor output as a claim and to admit that claim only when
fresh, structured, tracked-source-state-bound evidence satisfies a gate predicate. If the evidence
is admissible, the lifecycle advances; if it is missing, stale, incomplete, or outside the stated
trust model, the system repairs within a bounded loop, degrades honestly, escalates, or stops. In
the Proof-or-Stop implementation, this evidence is protected by authenticated integrity, producer
identity, and freshness checks over the exact tracked source state. We use ``proof''
operationally: a proof is gate-admissible evidence under that trust model, not a formal proof of
semantic program correctness. The broader engineering practice is \textbf{Loop Engineering}:
engineering the autonomous develop/review/test/done loop so that control decisions are made by
evidence-checking gates.

\paragraph{Semantic stance.} We separate the contribution into four layers. First, the semantic
shift is \emph{agent-as-claim}: an agent output may propose a lifecycle claim, but it is not itself
lifecycle state. Second, Proof-or-Stop Lifecycle Control is the methodology: lifecycle advancement
is a claim-admissibility decision. Third, evidence gates are the mechanism: claims advance only
when fresh, tracked-source-state-bound evidence satisfies the relevant predicate. Fourth,
receipts, source-state hashes, review runs, \textsc{done} gates, and the ablation, corpus, and
cross-vendor exhibits are the instantiation and evidence used to evaluate the method.

\begin{figure}[t]
\centering
\includegraphics[width=0.72\textwidth]{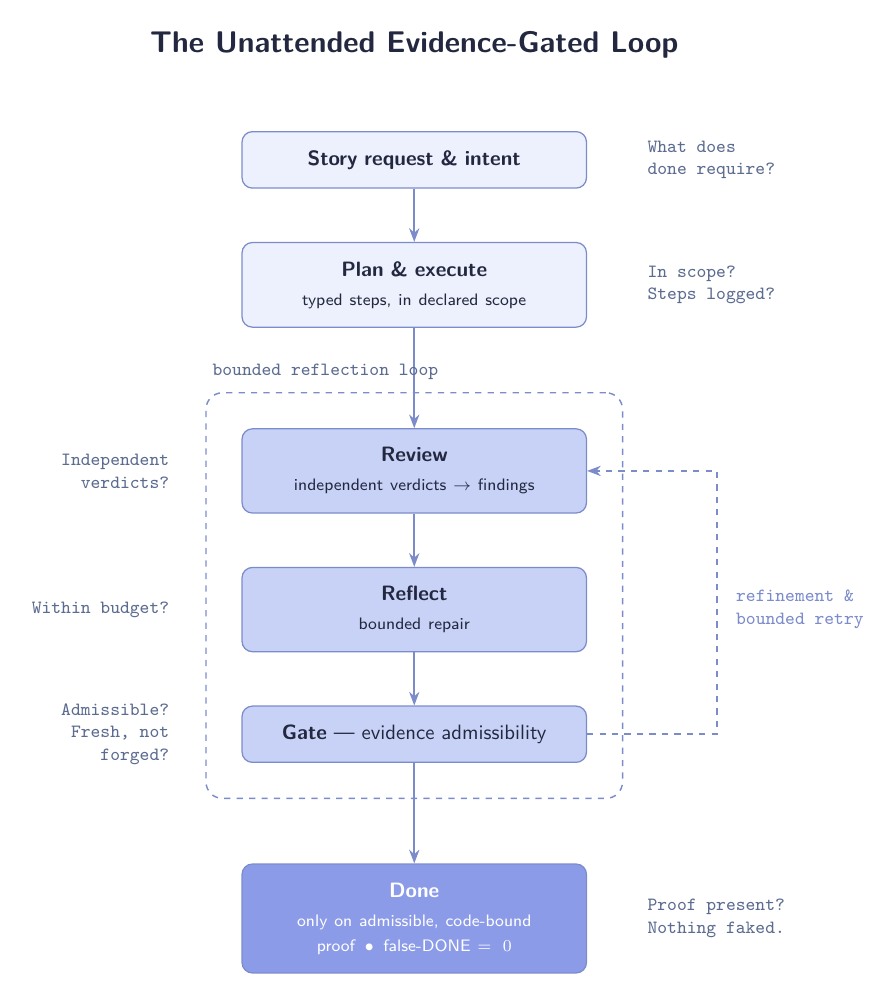}
\caption{Loop Engineering at a glance. The unattended develop loop runs
\textsc{plan}$\to$\textsc{execute}$\to$\textsc{review}$\to$bounded \textsc{reflect}$\to$\textsc{gate}$\to$\textsc{done};
every consequential transition is decided by an evidence gate (the monospace questions), not by
the agent. The loop advances only on admissible, code-bound evidence, loops back for a bounded retry
when evidence is short, and does not convert unsupported claims into \textsc{done}. One loop
iteration is shown in
technical detail in Fig.~\ref{fig:loop}. This is the proof-or-stop discipline: admissible
evidence advances the loop; missing, stale, or forged evidence blocks, repairs, or escalates
rather than advancing the lifecycle.}
\label{fig:loop-overview}
\end{figure}

\paragraph{Contributions.}
\begin{description}[leftmargin=1.4em,itemsep=2pt]
\item[\textnormal{\textbf{C1 --- Proof-or-Stop Lifecycle Control.}}] We define a
model-agnostic evidence-gated control method for autonomous coding lifecycles. Every
consequential lifecycle claim emitted by a host agent or workflow actor must reduce to
gate-consumable evidence before it can advance phase, review, test, \textsc{done}, or
merge-relevant state; ordinary notes and documentation remain light
(\S\ref{sec:spine}, \S\ref{sec:lifecycle}).
\item[\textnormal{\textbf{C2 --- Code-state-bound evidence mechanics.}}] We instantiate claim
admissibility through authenticated integrity, producer identity, and freshness binding. Evidence
carries fresh material, head, and story hashes; policy and command-set hashes; and a receipt
identity (command, arguments, working directory, exit code, output digest). Stale, reconfigured,
tampered, unauthorized, or build-proof-missing evidence is rejected
(\S\ref{sec:spine}, \S\ref{sec:lifecycle}).
\item[\textnormal{\textbf{C3 --- Host-neutral transfer as an evidence property.}}] We
characterize host-neutrality as local evidence over the same code identity rather than trust in a
remote host or protocol. Safety mechanisms and the cross-machine resume (HN-2) are \ver{}; the
strong cross-vendor quorum remains \gat{} (\S\ref{sec:hostneutral}).
\item[\textnormal{\textbf{C4 --- Operated evaluation and boundary discipline.}}] We report the
self-hosted Proof-or-Stop implementation and evidence from a 565-story / 1007-finding corpus,
powered ablation, injected-failure recovery, a 68-row cross-vendor high/critical review exhibit,
and prohibited-wording guard that limits unsupported capability claims
(\S\ref{sec:loop}, \S\ref{sec:recovery}, \S\ref{sec:selfapp}).
\end{description}

\paragraph{What is established, and what is not.} Table~\ref{tab:layered-evidence} separates the evidence
tiers used in the paper. The engine contract and evidence discipline are demonstrated in the
operated Proof-or-Stop corpus. The comparative ablation is a powered 9{,}240-cell result: the
primary A4-vs-A2$'$ contrast against the pre-registered budget-capped naive control has a
CI-excluding-zero not-amplification gain (\textbf{H1 $+1.6$pp, [0.8,2.5]}) and marginal
completion gain (H2). The
recovery claim is supported by both the powered full-matrix estimate (\S\ref{sec:recovery}) and
a verified pilot showing a loop-fidelity gradient on visible-test-passing wrong artifacts. Clean
tasks show overhead; green-but-wrong injected faults show risk-mitigation value, and the
68-row cross-vendor refresh shows that independent-vendor review caught high/critical defects in
the paper's own evidence machinery. The present evidence supports host-neutral gate semantics and
local receipt validation; strong cross-vendor
quorum claims remain gated on fresh independent host verdicts over the same material hash. These
boundaries are part of the method: claims are reported only at the tier supported by admissible
evidence. Cross-domain, multi-model, external-benchmark, and strong independent-host
generalization remain future work.

\paragraph{Roadmap.} Section~\ref{sec:background} states the claim-admissibility problem in
autonomous software work. Section~\ref{sec:spine} formalizes evidence admission, and
Section~\ref{sec:lifecycle} instantiates it in the Proof-or-Stop lifecycle. Sections~\ref{sec:loop}
and~\ref{sec:recovery} evaluate the unattended loop and injected-failure recovery.
Section~\ref{sec:hostneutral} covers host-neutral transfer; Section~\ref{sec:selfapp} reports
self-application and audit evidence; Section~\ref{sec:metrics} states the honest boundary of
experience reuse. Sections~\ref{sec:related}--\ref{sec:conclusion} cover related work, threats,
future work, and the conclusion.

\paragraph{Proof chain at a glance.} This paper does not ask the reader to accept one large
informal claim. It builds one chain: an agent output is a \emph{claim}; a lifecycle transition
needs \emph{admissible evidence}; a gate checks that evidence against the current tracked source
state; and missing or stale evidence blocks advancement. Table~\ref{tab:layered-evidence} is the
single map for current evidence, tier, and boundary. In the table, \emph{engine} evidence checks
that the gate behaves correctly; \emph{empirical} evidence tests whether acting on the gate helps against
weaker controls; \emph{independent-host} evidence is reserved for fresh multi-host receipts.

\begin{table}[H]
\centering\scriptsize
\caption{Proof chain, evidence tier, and current status for Proof-or-Stop Lifecycle Control.}
\label{tab:layered-evidence}
\begin{tabularx}{\textwidth}{@{}p{2.45cm}p{2.2cm}Y Y@{}}
\toprule
\textbf{Evidence source} & \textbf{Tier / status} & \textbf{What it supports} & \textbf{Boundary} \\
\midrule
Formal spine & engine / \ver{} &
Defines the agent-as-claim formal spine: actor outputs are lifecycle claims, and transitions
advance only when required claims are backed by admissible evidence
(Eqs.~\ref{eq:adm}--\ref{eq:advance}). &
A control method, not a semantic-correctness proof or a new coding model. \\
Receipt and tamper checks & engine / \ver{} &
Shows that gate inputs cannot be replaced by prose, stale logs, or tampered receipt bundles; 18
tamper classes rejected with zero false accepts / false rejects in the tested suite. &
Local-key trust model; does not defeat a compromised runner or prove claim semantics. \\
Engine contract & engine / \ver{} &
Shows \textsc{done}, review, test, and claim-boundary gates do not advance on self-report:
10/10 loop-engineering scenarios, false-\textsc{done}=0, plus a 10-group / 150-row
large-ledger stress suite. &
Checks gate behavior, not broad task success. \\
Powered ablation & empirical / \ver{} &
Tests whether enforcing the gate reduces visible-pass/hidden-fail amplification: A2$'$ $31/1800$
vs A4 $2/1800$, H1 not-amplified $+1.6$pp [0.8,2.5]. &
Rare-event, task-concentrated, one model family; powered ablation is control-policy evidence, not
full story-level receipt replay. \\
Cell03/Cell06 paired comparison & descriptive / \ver{} &
Shows terminal completion and Proof-or-Stop delivery admission can diverge on 1{,}152 matched cell
keys; 106 no-review completions were not admitted; the paired token-usage readout is 3.80$\times$
input+output tokens for the bundled gated run vs no-review control. &
Not hidden-oracle adjudicated; not an accuracy, dollar-cost, cost-benefit, isolated-overhead, or
completed multi-model result. \\
Recovery pilot & empirical / \ver{} &
Shows the mechanism path: bare loop amplifies wrong artifacts, one review gate safe-stops them,
bounded reflection repairs them. &
Pilot-sized, three tasks, B-fidelity proxy. \\
Self-application corpus & corpus / \ver{} &
Shows operated use and auditability: 565 stories / 1007 findings, 94.8\% resolved, 26/28 curated
deep-set findings filed while author tests were green. &
Self-built and selection-conditioned; not an unbiased population benchmark. \\
Cross-vendor exhibit & observational / \ver{} &
Shows independent-vendor review can catch load-bearing defects in same-vendor-passed artifacts:
68 high/critical Codex host-2 findings across 26 stories, all resolved. &
Existence and soundness exhibit; not a controlled marginal-rate estimate. \\
Host-neutral transfer & mechanism / \ver{} &
Shows local receipt and git-native handoff mechanics over the same tracked source identity. &
Strong cross-vendor quorum remains \gat{} until fresh independent-host receipts are powered. \\
Claim boundary and experience & boundary / \ver{} &
Guards unsupported capability wording; keeps runtime memory advisory with \code{gateEvidence:false}. &
Claim language and attention support only; not empirical superiority or causal learning. \\
PINN / Quantum bundles & future-domain smoke / \gat{} &
Preserved as schema smoke tests for future domain packages. &
Not current-result evidence; no PDE, quantum, solver, hardware, or scientific correctness claim. \\
\bottomrule
\end{tabularx}
\end{table}

\noindent\textbf{Terminology.} A \emph{claim} is an agent- or workflow-implied statement such as
reviewed, tested, \textsc{done}, or ready-to-merge. \emph{Admissible evidence} satisfies the
freshness, tracked-source-state binding, authenticated-integrity, producer-authorization, and
accepted-outcome checks in Eq.~\eqref{eq:adm}. \emph{Amplified} means a wrong artifact was shipped
or propagated; \emph{not-amplified} means it was repaired, stopped, or otherwise not propagated.

\begin{table}[t]
\centering\scriptsize
\caption{Operational evidence map for the claims above. The artifact workspace for this draft was extracted from the Proof-or-Stop experiment platform
checkpoint taken 2026-06-22 (Proof-or-Stop version 0.3.58).
Each row names the concrete evidence a reviewer can inspect, the command that produces or
summarizes it, and the gate or claim consumer that reads it.}
\label{tab:evidence-map}
\begin{tabularx}{\textwidth}{@{}p{3.0cm}Y p{3.25cm}Y@{}}
\toprule
\textbf{Claim} & \textbf{Artifact / evidence} & \textbf{Command} & \textbf{Gate consumer} \\
\midrule
\textsc{done} requires fresh full-test proof &
story \begin{tabular}[t]{@{}l@{}}\code{done-required-}\\\code{evidence.json}\end{tabular}:
command exits, output digests, \mh/\hh/\sfh,
policy/command-set hash &
\begin{tabular}[t]{@{}l@{}}\code{done\_required\_}\\\code{validate}\end{tabular} &
\textsc{test}$\to$\textsc{done}: rejects missing, stale, or command-set-drifted proof \\
Reviewer identity is structured, not prose &
\code{review-runs.json}: signed reviewer run, lane, round, \mh, reviewer identity &
\code{review\_run\_start} &
Review gate: pass/finding must reference a signed current-round, material/scope-fresh run \\
No-issue review is auditable &
\code{review-passes.json}: current-round pass with reviewer lane and run id &
\code{review\_pass\_set} &
\textsc{review}$\to$\textsc{test}: pass quorum and freshness floor \\
Findings block when severity/evidence require it &
\code{findings.json}: severity, category, status, evidenceState, resolution trail &
\code{finding\_add} &
\textsc{review}$\to$\textsc{test}: open verified critical/high findings block unless the
round-graded advisory rule applies \\
Unattended loop engine contract &
\code{final-report.json} artifacts aggregated into a loop-engineering suite report &
\code{baseline\_suite} &
Tier-A claim consumer: supports the 10/10 engine-contract result; not a lifecycle bypass \\
Stronger claim wording cannot be upgraded without evidence &
claim-boundary registry plus host-verdict / production-wording guard fields &
\code{quorum\_submit} + wording gate &
Tier-C claim consumer: stronger production or cross-host wording requires configured local
evidence; degraded fallback remains explicitly local \\
\bottomrule
\end{tabularx}
\end{table}

\section{Background and Problem}\label{sec:background}

Three lines of systems work have made autonomous agent work more robust, and each leaves the
trust problem open.

\emph{Durable / resumable execution} (Temporal-style workflows, graph runtimes, agent
frameworks with persistence, transactional agent runtimes) makes work \emph{survive} crashes
and restarts by checkpointing state. But surviving state is not verified state: resuming a
workflow that recorded ``tests passed'' re-asserts the recorded claim; it does not re-establish
that the claim matches the code now being merged.

\emph{Cross-vendor coordination} (agent-to-agent protocols, tool/context protocols, and the
broader family of agent-interoperability standards) \emph{routes messages and delegates tasks}
across heterogeneous systems. These are communication layers. They standardize how a verdict
travels, not whether it is admissible against the current artifact.

\emph{Evidence-driven release gates and deterministic verification loops} aggregate signals
into promote/hold/rollback decisions. This is the closest prior practice. The gap we address is
\emph{binding and scope}: a release gate typically makes one decision at the end, on coarse
signals; we bind \emph{every} lifecycle evidence item to code identity
(\mh/\hh/\sfh{}), add authenticated integrity digests and a receipt identity, and operate it lifecycle-wide
on a real self-hosted system.

The common failure mode these leave intact is that \textbf{a green pipeline can coexist with a
real defect}. If downstream automation treats that green status as sufficient evidence, it may
advance, merge, and mark the work done. Our deep finding set (\S\ref{sec:selfapp}) measures how
often ``author's tests pass'' failed to imply ``correct'' in the operated corpus.

The agent setting makes this sharper than ordinary CI slippage. A conventional gate can often
assume the actor is honest but fallible; an unattended coding agent can generate the code, retry
until a visible check turns green, and narrate or package success. The gate must therefore verify
state-bound evidence rather than trust the actor's report. This is not only a theoretical risk: in
the powered ablation, the compute-budgeted naive loop amplified $31/1800$ visible-pass/hidden-fail
injected cells while the gated loop amplified $2/1800$ (Table~\ref{tab:powered-ablation}); in the
self-application deep set, 26 of 28 findings were filed while the author's tests were green
(\S\ref{sec:selfapp}).

\begin{table}[t]
\centering\scriptsize
\caption{Why existing gates are not enough for agent lifecycle claims. These mechanisms compose
with Proof-or-Stop, but none by itself decides whether a current \emph{reviewed/tested/done/ready}
claim is admissible for the exact source state about to advance.}
\label{tab:why-not-existing-gates}
\begin{tabularx}{\textwidth}{@{}p{2.8cm}p{2.7cm}p{2.7cm}Y@{}}
\toprule
\textbf{Mechanism} & \textbf{Control unit} & \textbf{Binding/freshness} & \textbf{Gap for autonomous-agent work} \\
\midrule
CI/CD gate & Job or pipeline result & Usually the commit/job that ran; often consumed as a green status &
Answers ``did this configured job pass?'', not whether the lifecycle claim is complete, current,
authorized, and sufficient to advance review/test/\textsc{done}/merge. \\
SLSA / in-toto provenance & Artifact production history & Strong provenance/attestation over build or supply-chain steps &
Useful evidence source, but it attests production history rather than deciding every agent
lifecycle transition; a source-state drift or missing reviewer/test claim still needs a gate. \\
Durable execution & Persisted workflow state & Resumes recorded state after interruption &
Preserves work, including stale or false claims; it does not re-establish that ``tests passed'' or
``reviewed'' still supports the current source state. \\
A2A / MCP / agent frameworks & Messages, tools, delegation, agent orchestration & Transports context or routes work between agents/tools &
Moves claims and verdicts, but does not decide whether those claims are admissible evidence against
the artifact being advanced. \\
Proof-or-Stop & Lifecycle claim transition & Fresh tracked-source-state digest, receipt identity,
policy/command-set digest, accepted outcome & Converts each lifecycle-moving claim into an
evidence-admission decision: advance only on admissible evidence; otherwise repair, degrade,
escalate, or stop. \\
\bottomrule
\end{tabularx}
\end{table}

\begin{table}[t]
\centering\scriptsize
\caption{Where Proof-or-Stop sits relative to ordinary coding-agent use. Product names are
illustrative host examples; the comparison is between assurance layers, not between vendor feature
sets. Proof-or-Stop composes with host agents and CI rather than replacing them.}
\label{tab:modes}
\begin{tabularx}{\textwidth}{@{}p{2.25cm}YYY@{}}
\toprule
\textbf{Dimension} & \textbf{Bare host-agent session} &
\textbf{Host agent + CI/CD} & \textbf{Host agent + Proof-or-Stop} \\
\midrule
Semantic model & Agent report may be consumed as lifecycle state. & Job status is consumed as a
lifecycle signal. & Agent output is a claim; only admissible evidence advances state. \\
Primary role & Generate, edit, run tools, and explain work. & Generate work, then run configured checks. &
Generate work, then admit lifecycle claims only through evidence gates. \\
Review / test trust & Agent or author can narrate reviewed/tested/done. & Green job shows configured
checks passed. & Reviewer verdicts and test receipts must be fresh, structured, and source-state-bound. \\
Failure mode & Plausible self-report or visible-check overfitting can advance. & Green pipeline can
coexist with hidden failure or stale evidence. & Missing, stale, forged, or incomplete evidence blocks,
repairs, degrades, escalates, or stops. \\
Audit trail & Conversation, files, and git history. & Job logs plus git history. & Receipt identity,
material/source digests, policy and command-set digests, review runs, findings, and \textsc{done}
certificates. \\
Merge/\textsc{done} decision & Human or host judgement. & Branch policy / CI status. & Current
tracked-source-state-bound certificate consumed by lifecycle gates. \\
\bottomrule
\end{tabularx}
\end{table}

\section{The Evidence-Gating Principle}\label{sec:spine}

\paragraph{Method abstraction.} Proof-or-Stop Lifecycle Control reduces lifecycle advancement
to a claim-admissibility decision. The core method is:
\[
\boxed{
\begin{gathered}
\textbf{Proof-or-Stop}
=
\text{Actor output}\ \longrightarrow\ \text{Claim}\ \longrightarrow\ \text{Evidence}\ \longrightarrow\ \text{Gate}\\
\longrightarrow\ \text{Lifecycle transition}.
\end{gathered}}
\]
Proof-or-Stop controls each lifecycle transition; repeated proof-or-stop transitions form the
bounded lifecycle loop. In the autonomous-coding implementation evaluated here, actors include
agents, reviewers, tools, and workflow commands. Their outputs are interpreted as lifecycle
claims; claims require admissible evidence; evidence is checked by gate predicates; and only
passing gates may advance lifecycle state. Evidence-gated claim admissibility is the enforcement
mechanism for this method.
This is the operational form of \emph{agent-as-claim} semantics: the actor may emit the claim, but
the lifecycle transition is decided by evidence admission.

The control abstraction is domain-neutral: actor output is treated as a claim, claims require
admissible evidence, and gates decide whether a lifecycle transition may advance. This paper
empirically evaluates the abstraction in autonomous coding lifecycles. Cross-domain
instantiations require domain-specific evidence packages and separate evaluation.

We make the thesis precise. Let $H$ be the current tracked source state of a unit of work, identified by
content digests over the version-control tree. We define three identities, computed over
\code{git ls-tree} with the lifecycle's own metadata excluded so that recording evidence does
not perturb the hash it is bound to:
\begin{align}
\mh(H) &= \mathrm{SHA256}\big(\mathrm{canon}(\text{tracked source tree at }H,\ \setminus\ \text{metadata})\big), \label{eq:mh}\\
\hh(H) &= \text{commit identity of }H, \nonumber\\
\sfh(H) &= \mathrm{SHA256}\big(\mathrm{canon}(\text{story-owned files at }H)\big). \nonumber
\end{align}
These identities bind the tracked source state. They are not, by themselves, a full executable-state
attestation: dependency resolution, toolchain/container/OS, environment variables, external service
state, and untracked generated files must be covered by policy, command, environment, or dependency
digests if a deployment needs that stronger claim.

A piece of \emph{evidence} $E$ is a structured record (not prose) describing the outcome of a
checkable action: a test run, a build, a scope check, a reviewer verdict. Evidence carries a
\emph{binding} $\beta(E) = \langle \mh_E,\hh_E,\sfh_E,\textsf{policyHash}_E,\textsf{commandSetHash}_E\rangle$
and, for executed actions, a \emph{receipt identity}
$\rho(E)=\langle\text{cmd},\text{args},\text{cwd},\text{exit},\textsf{outputDigest}\rangle$.
Signed local receipts and review verdicts also carry producer identity, such as actor, lane, host
or session, and signing-key identity.

\paragraph{Admissibility.} A gate admits evidence $E$ for claim $c$ at state $H$ iff
\begin{equation}
\resizebox{\linewidth}{!}{$\displaystyle
\begin{aligned}
\mathrm{Admissible}(E,c,H) \;\equiv\;&
\mathrm{Fresh}(E,H)\ \wedge\ \mathrm{Complete}(E)\ \wedge\
\mathrm{IntegrityVerified}(E)\ \wedge\ \mathrm{ProducerAuthorized}(E)\\
&{}\wedge\ \mathrm{ExecutionAttested}(E)\ \wedge\
\mathrm{Supports}(E,c)\ \wedge\ \mathrm{OutcomeAccepted}(E).
\end{aligned}
$}.
\label{eq:adm}
\end{equation}
Here $\mathrm{Fresh}$ checks the required \mh/\hh/\sfh{} bindings against the live state $H$;
$\mathrm{Complete}$ checks required policy and command-set bindings; $\mathrm{IntegrityVerified}$
checks signatures and digest chains; $\mathrm{ProducerAuthorized}$ checks that the actor, lane,
host/session, or signing key is authorized for the claim; $\mathrm{ExecutionAttested}$ checks command,
arguments, working directory, exit code, and output digest when execution is required;
$\mathrm{Supports}$ ties the record to claim $c$; and $\mathrm{OutcomeAccepted}$ checks the
gate-specific pass or accepted-degradation outcome. The freshness conjunct defeats the most common
silent failure: evidence produced against an earlier source state is rejected the instant the source
tree changes, because $\mh_E \neq \mh(H)$ (Eq.~\eqref{eq:mh}). Missing build proof, a reconfigured
command set ($\textsf{commandSetHash}$ mismatch), an unauthorized producer, a failed execution
attestation, or a hand-edited metadata file all fail Eq.~\eqref{eq:adm}. Local-key receipts provide
authenticated integrity, producer identity, and freshness under the stated single-host trust
assumption; they do not independently prove execution truth against a compromised runner or semantic
correctness of the claim.

\paragraph{Offline receipt-bundle contract.} B-4c65 exercises Eq.~\eqref{eq:adm} as an
adversarial contract test for the ``lying agent'' case: an agent may claim that DONE evidence is
fresh, signed, and passing, but the verifier accepts only a self-contained bundle whose binding
and receipt identity re-check offline. The implementation exposes a programmatic API that builds
a local-key-signed bundle from DONE-story evidence and verifies it without network or server
access. Its assurance label is deliberately \code{local-key-single-host}: it proves portable
freshness and tamper-evidence for a local receipt, not an independent multi-host quorum. We call this
\emph{local-key scope} below.

\begin{table}[H]
\centering\scriptsize
\caption{Offline receipt-bundle adversarial contract (B-4c65). The authentic bundle is accepted;
18 tamper classes are rejected; all 10 non-\code{ok} reason codes are reached; false-accept $=0$
and false-reject $=0$. This is local-key scope, not an independent multi-host quorum.}
\label{tab:receipt-bundle-contract}
\begin{tabularx}{\textwidth}{@{}>{\raggedright\arraybackslash}p{2.6cm}
Y
>{\raggedright\arraybackslash}p{3.1cm}
>{\raggedright\arraybackslash}p{1.55cm}@{}}
\toprule
\textbf{Case family} & \textbf{Examples} & \textbf{Verifier reason} & \textbf{Result} \\
\midrule
Authentic control &
Untampered signed bundle over matching material hash and trusted public key &
\code{ok} &
accepted \\
Freshness / binding drift &
Code-byte flip; stale material hash; incomplete binding &
\begin{tabular}[t]{@{}l@{}}\code{stale};\\\code{binding\_incomplete}\end{tabular} &
rejected \\
Signature tamper &
Evidence edit; host-verdict swap; command-set change; forged signature &
\code{signature\_invalid} &
rejected \\
Missing / wrong signer &
Empty signature; signer public key differs from the trusted key &
\begin{tabular}[t]{@{}l@{}}\code{signature\_missing};\\\code{signer\_key\_mismatch}\end{tabular} &
rejected \\
Malformed structure &
Null host verdict; null decision; extra top-level key &
\code{malformed\_bundle} &
rejected \\
Receipt identity malformed &
Non-SHA256 output digest &
\code{digest\_mismatch} &
rejected \\
Receipt did not pass &
Receipt exitCode 1; exitCode -1; empty command &
\begin{tabular}[t]{@{}l@{}}\code{receipt\_not\_}\\\code{passing}\end{tabular} &
rejected \\
Missing proof / bad decision &
No receipts; signed bundle with decision \code{fail} &
\begin{tabular}[t]{@{}l@{}}\code{build\_proof\_}\\\code{missing};\\\code{decision\_not\_pass}\end{tabular} &
rejected \\
\bottomrule
\end{tabularx}
\end{table}

\paragraph{Gate-strength self-red-team contract.} A verifier is only as good as its
discrimination: a gate that blocks everything has a perfect ``catch rate'' yet ships nothing.
B-7b6e instruments this directly by scoring a reviewer over a hidden-ground-truth corpus of
\emph{passes-the-visible-test-but-wrong} artifacts plus one genuine control. The score is the pair
$(\textsf{catchRate}=\textsf{caught}/\textsf{wrongTotal},\ \textsf{falseBlock})$, because catch
rate alone is gameable. Three \emph{reference} reviewers make the metric non-vacuous:
a test-only reviewer misses every wrong-but-test-passing artifact; a block-all reviewer reaches
$\textsf{catchRate}=1.0$ but falsely blocks the genuine control; and an oracle reviewer reaches
$\textsf{catchRate}=1.0$, $\textsf{falseBlock}=0$, dominating both. The score is itself evidence:
B-7b6e publishes it as a B-4c65 receipt bundle in local-key scope, and the
gate-strength verifier re-derives the result digest so the receipt binds not just ``a pass
happened'' but ``this discrimination score, over this hidden-truth corpus, is the one attested.''
Honest scope: these are reference endpoints and an oracle, not a live review lane; integrating the
metric into a live artifact-reading review lane remains explicit future work.

\begin{table}[H]
\centering\scriptsize
\caption{Gate-strength self-red-team contract (B-7b6e), 12 hermetic checks, all passing.
The three reference reviewers bracket the achievable space, proving catch rate alone is gameable
and that the published metric is offline-verifiable and tamper-evident (including against a result
tampered \emph{after} bundling). Local-key scope; reference reviewers, not a live
review lane.}
\label{tab:gate-strength-contract}
\begin{tabularx}{\textwidth}{@{}>{\raggedright\arraybackslash}p{3.2cm}
>{\raggedright\arraybackslash}X
>{\raggedright\arraybackslash}p{3.0cm}@{}}
\toprule
\textbf{Contract check} & \textbf{What it proves} & \textbf{Outcome} \\
\midrule
test-only reviewer & ships every wrong-but-test-passing probe & $\textsf{catchRate}=0$ \\
block-all reviewer & catch rate alone is gameable & $\textsf{catchRate}=1.0$, $\textsf{falseBlock}=\textsf{genuineTotal}$ \\
oracle dominance & real discrimination needs both axes & out-catches test-only \emph{and} under-blocks block-all \\
empty / all-genuine corpus & no vacuous $0/0=1.0$ & rejected \\
invariant-violating corpus & oracle is well-defined & rejected \\
receipt verifies offline & metric is portable evidence & accepted (local key) \\
tampered signed field & bundle is tamper-evident & rejected (bad signature) \\
stale materialHash & freshness binding holds & rejected (\code{stale}) \\
authentic result match & result--receipt loop closed & matches \\
tampered result (catchRate / reason) & full result is bound & does not match \\
\bottomrule
\end{tabularx}
\end{table}

\paragraph{Evidence-instrument characterization contract.} An evidence gate is only as
trustworthy as the instruments that produce its evidence. Where B-4c65 and B-7b6e red-team a
\emph{receipt} and a \emph{reviewer}, the proof-or-stop replay campaign (five DONE stories
B-9e44\,/\,B-bae9\,/\,B-2a5b\,/\,B-f511\,/\,B-853a) red-teams a \emph{producer}: the
memory/playbook runtime baseline verifier. Against a frozen pre-registered protocol
(\code{protocol.json}, \textsf{sha256} \code{a3a781fed2\ldots}) it characterizes the instrument on
four properties --- \emph{replay-determinism}, \emph{tamper-fail-closed}, \emph{resume-integrity},
and \emph{scale-envelope} --- and adjudicates them with a purely offline three-state
(pass\,/\,fail\,/\,inconclusive) analyzer that reads only committed evidence and re-verifies every
ledger record's digest before adjudicating. The resume-integrity arm is the proof-or-stop discipline
itself under injected interruption: boundary and mid-iteration kills leave, on resume, exactly one
identity-keyed record per iteration with zero duplicates and zero gaps. The analyzer emits a report
whose digest \code{198e33866b\ldots} $= \textsf{sha256}(\textsf{canonicalJson}(\textit{report} -
\textit{digest}))$ was re-derived byte-for-byte offline and \emph{cross-vendor} (an independent
Codex host recomputed it), and a hermetic 19-check selfcheck proves the three-state logic
un-gameable: incomplete, stub, mislabeled, non-physical, or tampered evidence each yields
fail\,/\,inconclusive\,/\,abort, never a false all-pass, and the claim text is emitted only when all
four properties hold. Honest scope: this characterizes one instrument under a frozen protocol over
enumerated scenarios; the report digest is a reproducibility and tamper-evidence anchor at
local\,+\,cross-vendor-replication scope, not an authenticated multi-host quorum, and the study is
not Tier-C (a signing adversary who recomputes digests is out of scope).

\begin{table}[H]
\centering\scriptsize
\caption{Evidence-instrument characterization contract (proof-or-stop replay campaign, M0--M4). Four
properties of the baseline verifier, adjudicated offline to a three-state verdict; all pass under the
frozen protocol over the enumerated scenarios. The report digest is re-derived cross-vendor; a
19-check hermetic selfcheck proves no incomplete/stub/mislabeled/tampered evidence yields a false
all-pass. Frozen-protocol, enumerated scope; local\,+\,cross-vendor-replication, not a multi-host
quorum; not Tier-C.}
\label{tab:instrument-characterization}
\begin{tabularx}{\textwidth}{@{}>{\raggedright\arraybackslash}p{2.7cm}
Y
>{\raggedright\arraybackslash}p{1.4cm}@{}}
\toprule
\textbf{Property} & \textbf{Evidence (enumerated, frozen protocol)} & \textbf{Verdict} \\
\midrule
replay-determinism &
33/33 proof records (30 small + 3 full $10^6$-row) pairwise-identical on digests and gold metrics, at both scales &
pass \\
tamper-fail-closed &
27/27 (class $\times$ handling) tamper verdicts conform (18 fail-closed, 2 accepted, 6 acknowledged-limitation, 1 dropped); 0 non-conforming &
pass \\
resume-integrity (proof-or-stop) &
boundary + mid-iteration kills $\to$ on resume exactly one record per iteration, 0 duplicate / 0 gap, casualty resolved to stop-or-clean-rerun &
pass \\
scale-envelope &
every verify within the 128.2\,s budget cap; full-profile ($10^6$ rows) max 18.7\,s (14.6\% of cap) &
pass \\
\bottomrule
\end{tabularx}
\end{table}

\begin{figure}[t]
\centering
\resizebox{\textwidth}{!}{%
\begin{tikzpicture}[
  node distance=6mm,
  box/.style={draw=pbEdge,rounded corners=2pt,align=center,inner sep=4pt,font=\small,fill=pbPale},
  gate/.style={draw=pbDeep,rounded corners=2pt,align=center,inner sep=5pt,font=\small\bfseries,fill=pbDeep,text=white,thick},
  ev/.style={draw=pbEdge,rounded corners=2pt,align=center,inner sep=4pt,font=\scriptsize,fill=pbPale},
  >=Stealth]
\node[box] (agent) {Agent\\(produces work)};
\node[ev,right=14mm of agent] (evid) {Evidence $E$\\\scriptsize cmd,args,cwd,exit,\\digest $+$ \mh/\hh/\sfh\\$+$ policy/cmdset hash};
\node[gate,right=14mm of evid] (gate) {GATE\\check Eq.\,(2)};
\node[box,above right=4mm and 14mm of gate] (adv) {Advance\\$\phi_i\!\to\!\phi_{i+1}$};
\node[box,below right=4mm and 14mm of gate,fill=pbMid,draw=pbDeep] (blk) {Block /\\fail-closed};
\draw[->] (agent) -- (evid);
\draw[->] (evid) -- (gate);
\draw[->] (gate) -- node[above,font=\scriptsize]{Admissible} (adv);
\draw[->] (gate) -- node[below,font=\scriptsize]{stale/forged/missing} (blk);
\draw[->,dashed,pbEdge] (agent.south) to[out=-30,in=180] node[below,font=\scriptsize\itshape]{self-report: not gate evidence} ($(gate.south west)+(-2mm,-9mm)$);
\end{tikzpicture}}
\caption{The evidence-gating spine. The agent's work is reduced to structured evidence bound to
code identity; the gate decides the transition by checking Eq.~\eqref{eq:adm}. Self-report
(dashed) is not admitted as gate evidence.}
\label{fig:gate}
\end{figure}
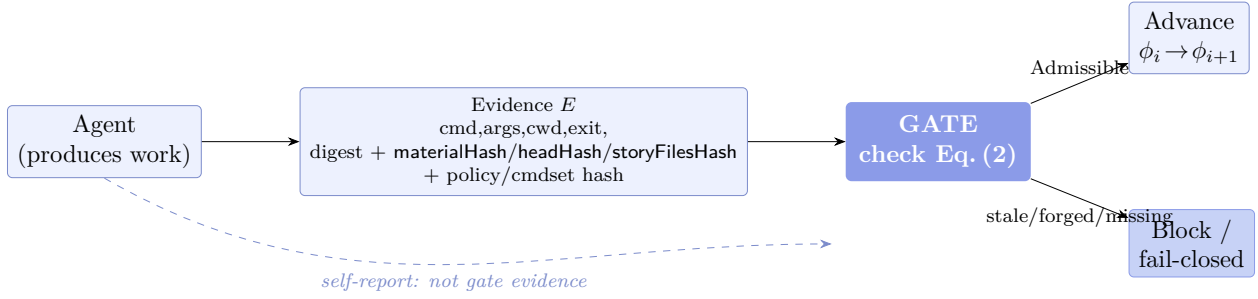

\paragraph{Gated advancement.} Let a unit of work pass through phases
$\phi_1 \to \phi_2 \to \cdots$. Each transition $\phi_i \to \phi_{i+1}$ declares a set of
\emph{required claims} $C_{i}$ (e.g.\ ``scope is frozen,'' ``tests pass,'' ``review
quorum met''). The transition is permitted iff every required claim is backed by admissible
evidence:
\begin{equation}
\mathrm{Advance}(\phi_i \to \phi_{i+1}, H)\ \Longleftrightarrow\
\forall\, c \in C_i\ \ \exists\, E_c\ :\ \mathrm{Provides}(E_c, c)\ \wedge\ \mathrm{Admissible}(E_c, c, H).
\label{eq:advance}
\end{equation}
Crucially, a natural-language report from an agent is not an $E_c$: it provides neither verified
integrity nor attested execution and carries no binding. Eq.~\eqref{eq:advance} therefore has no term for
self-report (Fig.~\ref{fig:gate}).

\paragraph{Scope (why this is not too heavyweight).} The heavy machinery of
Eqs.~\eqref{eq:adm}--\eqref{eq:advance} applies \emph{only} to claims in $\bigcup_i C_i$ ---
those that move phase, pass review, certify tests, mark done, or merge. Ordinary developer
notes, design rationale, and documentation are \emph{advisory}: they inform attention, never a
gate, and are deliberately excluded from every binding (\S\ref{sec:metrics}). The discipline is
expensive exactly where being wrong is expensive, and cheap everywhere else.

\section{Lifecycle Instantiation}\label{sec:lifecycle}

Proof-or-Stop instantiates Eqs.~\eqref{eq:adm}--\eqref{eq:advance} as a command-line lifecycle for a
unit of work it calls a \emph{story}. A story moves
\textsc{init}\,$\to$\,\textsc{init-check}\,$\to$\,\textsc{plan}\,$\to$\,\textsc{dev}\,$\to$\,\textsc{review}\,$\to$\,\textsc{test}\,$\to$\,\textsc{done},
and each arrow is an instance of Eq.~\eqref{eq:advance} with concrete required claims:

\begin{itemize}[leftmargin=1.4em,itemsep=1pt]
\item \textsc{plan}$\to$\textsc{dev}: an adversarial plan review must persist a structured
result (not a ``looks good''); scope is frozen into a contract bound to declared file paths.
\item \textsc{dev}$\to$\textsc{review}: a scope-contract check verifies the diff touches only
contracted paths; out-of-scope edits fail closed.
\item \textsc{review}$\to$\textsc{test}: independent reviewer lanes must persist verdicts;
critical/high findings block; a per-run test receipt must be admissible.
\item \textsc{test}$\to$\textsc{done}: see the full-test receipt and review-assurance floor
below.
\end{itemize}

\begin{table}[t]
\centering\scriptsize
\caption{Lifecycle gate matrix. Each transition names the structured artifact, the gate check, and
the fail-closed behavior when proof is missing or stale.}
\label{tab:lifecycle-gates}
\begin{tabularx}{\textwidth}{@{}p{2.05cm}Y Y Y@{}}
\toprule
\textbf{Transition} & \textbf{Evidence artifact} & \textbf{Gate check} &
\textbf{Fail-closed behavior} \\
\midrule
\textsc{plan}$\to$\textsc{dev} & structured plan review / story contract &
scope exists and story files hash is current & refuse DEV until plan evidence exists \\
\textsc{dev}$\to$\textsc{review} & scope-contract check + allowed paths &
diff is within declared scope and bound to current \mh{} & refuse REVIEW on out-of-scope edits \\
\textsc{review}$\to$\textsc{test} &
\begin{tabular}[t]{@{}l@{}}\code{review-runs.json}\\\code{review-passes.json}\\\code{findings.json}\end{tabular} &
current round, reviewer identity, material/scope freshness; pass may be non-latest &
block TEST on stale/scope-drifted pass or open verified high/critical finding \\
\textsc{test}$\to$\textsc{done} &
\begin{tabular}[t]{@{}l@{}}\code{done-required-}\\\code{evidence.json}\end{tabular} &
tree hashes plus policy/command-set hash all match current tree &
block DONE on stale, missing, or command-set-drifted proof \\
high-risk DONE & host verdict receipts / local review-assurance status &
3$\times$2 independent verdicts over current \mh{}, or explicit degraded fallback &
degrade honestly; do not upgrade local fallback into a stronger claim \\
\bottomrule
\end{tabularx}
\end{table}

\begin{figure}[t]
\centering
\resizebox{0.94\textwidth}{!}{%
\begin{tikzpicture}[
  node distance=9mm,
  flowbox/.style={draw=pbEdge,rounded corners=2pt,align=center,inner sep=5pt,font=\small,fill=pbPale},
  gate/.style={draw=pbDeep,rounded corners=2pt,align=center,inner sep=5pt,font=\small\bfseries,fill=pbDeep,text=white,thick},
  >=Stealth]
\node[flowbox] (cmd) {Command\\\scriptsize e.g. \code{review\_run\_start}\\\code{done\_required\_validate}};
\node[flowbox,right=of cmd] (artifact) {Structured artifact\\\scriptsize JSON receipt / verdict\\not prose};
\node[flowbox,right=of artifact] (bind) {Binding\\\scriptsize \mh/\hh/\sfh{}\\policy + command hashes};
\node[gate,right=of bind] (consumer) {Gate consumer\\\scriptsize REVIEW / TEST / DONE};
\node[flowbox,above right=5mm and 9mm of consumer] (advance) {advance};
\node[flowbox,below right=5mm and 9mm of consumer,draw=pbDeep,fill=pbMid] (block) {block / degrade};
\draw[->] (cmd) -- (artifact);
\draw[->] (artifact) -- (bind);
\draw[->] (bind) -- (consumer);
\draw[->] (consumer) -- node[above,font=\scriptsize]{admissible} (advance);
\draw[->] (consumer) -- node[below,font=\scriptsize]{missing, stale, forged} (block);
\end{tikzpicture}}
\caption{Operational evidence flow. Commands produce structured artifacts, artifacts carry code
and policy bindings, and gate consumers decide whether to advance, block, or honestly degrade.}
\label{fig:evidence-flow}
\end{figure}
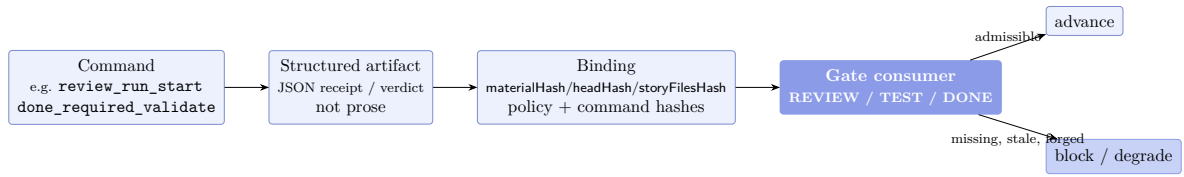

\paragraph{Review-run binding race.} B-6c4d refined the REVIEW evidence rule after a
concurrency failure in which a reviewer retrying \code{review\_run\_start} minted a newer
\code{reviewRunId} on the same lane and round, causing an earlier no-issue pass to be rejected
solely by id-recency even though it was still bound to the same material and scope. Id-recency is
therefore not itself a freshness predicate. A PASS is admissible from any signed run in the
current review round if the run's \mh/\sfh{} and scope binding still match the live story state.
The relaxation is kind-aware: open FINDING evidence still requires the lane's latest run, so a
newer pass cannot hide an unresolved older finding. The only whole-tree-drift downgrade through
\sfh{} also excludes \code{EMPTY\_STORY\_FILES\_HASH} on both the recorded and live side; the hash
of an empty story-owned diff carries no story-owned signal and falls back to strict \mh{}
freshness. Thus the gate removes a false rejection without weakening the surviving round, lane,
signature, freshness, and scope checks.

\paragraph{The \textsc{done}-required full-test receipt.} Before a story may enter
\textsc{done}, Proof-or-Stop requires a freshness-bound, authenticated integrity receipt that a configured set of
commands (a build plus the full test suite) ran to success \emph{at the current code state}.
The receipt records each command, its exit code, an output digest, the \mh/\hh/\sfh{}, and a
$\textsf{policyHash}/\textsf{commandSetHash}$; it is rejected if any command set was altered or
if the source tree has drifted since. This is C2 in its sharpest form: ``done'' is not the
agent saying done, and not even a green test log --- it is a receipt that re-derives, bound to
the exact tree being merged.

\paragraph{The review-assurance floor.} For material code changes on high-risk paths, a
\textsc{done} transition additionally requires a multi-round, multi-host review floor. Let a
\emph{host verdict} be admissible (Eq.~\eqref{eq:adm}) and let two verdicts be \emph{independent}
if they differ in host, session, and signing-key identity. Let $S$ denote the current tracked source
state for this floor. The full floor is
\begin{equation}
	\mathrm{Floor}_{R\times K}(S) \;\equiv\;
	\big|\{\, r : |\mathrm{IndepPass}(r,S)| \ge K \,\}\big| \;\ge\; R,
	\qquad (R,K)=(3,2),
\label{eq:floor}
\end{equation}
i.e.\ at least $R{=}3$ rounds each carrying $\ge K{=}2$ independent passing verdicts over the
\emph{current} \mh. When a second host is genuinely unavailable, the system does not synthesize a
quorum: it records a \emph{degraded single-host} fallback that remains local-only, preserving
honesty by construction:
\begin{equation}
\resizebox{\linewidth}{!}{$\displaystyle
\mathrm{LocalAssurance} \;=\; \mathrm{FullQuorum} \vee \mathrm{ExplicitDegradedFallback},
\qquad
\mathrm{FullAssurance} \;=\; \mathrm{FullQuorum},
\qquad \text{degraded} \Rightarrow \mathrm{FullAssurance}=\textsf{false}
$}.
\label{eq:localfloor}
\end{equation}

\paragraph{Pull requests as interface, not trust boundary.} The 3$\times$2 host
\textsc{done} floor shifts the trust root for merge-readiness away from a pull-request comment
thread or approval checkbox and toward a Proof-or-Stop evidence certificate. A pull request may
remain a useful display, discussion, and audit surface, but the admissibility decision is the
current-tracked-source-state-bound evidence bundle: fresh full-test receipts, admissible review verdicts,
material-hash freshness, and the required review-assurance floor. In the full-quorum case, this
supports a \textsc{done} or merge-readiness certificate that a repository could consume as a merge
gate. In the degraded single-host case, the certificate remains local assurance only and must not
be upgraded into full merge-readiness. This paper therefore supports PR-independent
merge-readiness as a mechanism claim; it does not claim general PR-less auto-merge or production
release.

A merge consumer must verify the certificate against the exact source commit it is about to merge
and perform the merge under a protected compare-and-swap condition: if either the source commit or
the target branch head changes between verification and merge, the certificate is stale and the
merge is refused.

We observed Eqs.~\eqref{eq:adm} and \eqref{eq:floor} act in practice while preparing this very
work: a small fix re-merged the main line under it, which advanced the tree hash and
\emph{invalidated} previously-submitted host verdicts (their $\mh_E$ no longer matched $\mh(H)$);
the floor refused \textsc{done} until the verdicts were re-attested over the new \mh{} --- the
freshness conjunct of Eq.~\eqref{eq:adm} doing exactly its job.

\section{The Unattended Loop}\label{sec:loop}

\begin{figure}[t]
\centering
\includegraphics[width=\textwidth]{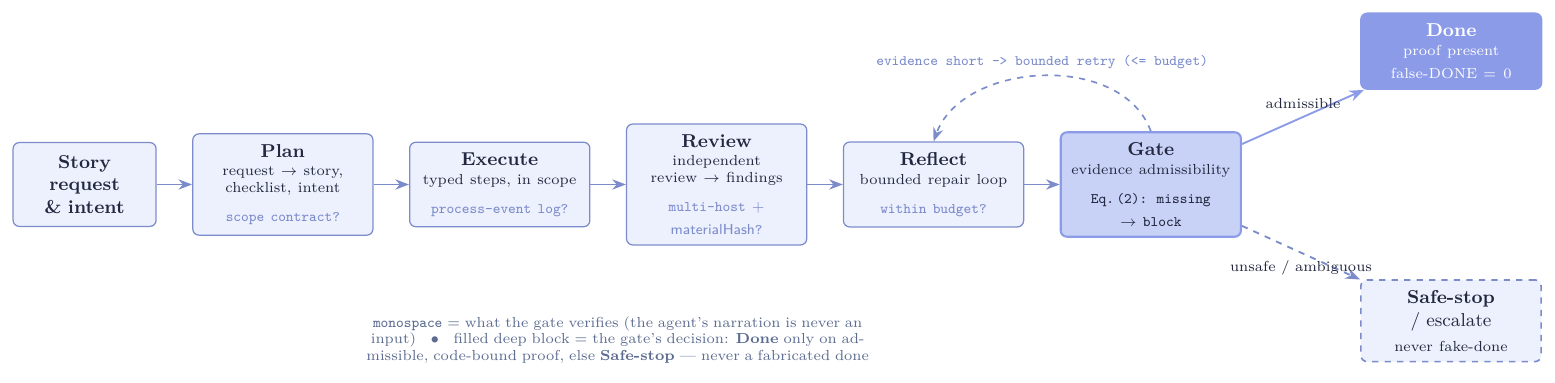}
\caption{The unattended evidence-gated loop. Each stage couples \emph{what the agent does} (black)
with \emph{the evidence the gate verifies} (green); the gate advances only on admissible,
code-bound proof (Eq.~\eqref{eq:adm}), loops back for a bounded retry when evidence is short, and
fails closed to safe-stop or escalation rather than unsupported \textsc{done} --- the engine
contract of Table~\ref{tab:udm} ($0$ false-DONE).}
\label{fig:loop}
\end{figure}

To evaluate the engine contract, we run Proof-or-Stop as a single evidence-gated loop
(Figure~\ref{fig:loop}):
\textsc{plan}$\to$\textsc{execute}$\to$\textsc{review}$\to$bounded
\textsc{reflect}$\to$\textsc{gate}$\to$\textsc{done}. At every step, agent actions are reduced
to structured, code-bound evidence; the gate decides whether to advance, loop back for a bounded
retry, or stop safely. The machinery is first verified as a contract (\S\ref{sec:loopA}) and
then compared with weaker control regimes
(\S\ref{sec:loopB}).
This is proof-or-stop control: the loop may continue only by producing admissible evidence, not
by repeating or rephrasing a lifecycle claim.

\subsection{Engine contract (verified)}\label{sec:loopA}

To evaluate the loop contract, we encode a full unattended develop loop as ten scenarios
(Table~\ref{tab:udm}), each asserting whether the engine should advance or block. The suite
passes 10/10 with zero false-DONE and a safety success rate of $1.0$ on the engine harness. This
is a Tier-A correctness check of the machinery: it demonstrates that the gate enforces the
contract, but it does not by itself show that acting on the contract improves outcomes against a
baseline. That comparison is \S\ref{sec:loopB}.

\begin{table}[t]
\centering\small
\caption{Unattended-loop engine contract (Tier A, \ver{}: 10/10, false-DONE $=0$).
``Expected'' \textsc{block} means the scenario must \emph{refuse} to advance.}
\label{tab:udm}
\begin{tabular}{@{}llc@{}}
\toprule
\textbf{Scenario} & \textbf{The engine asserts} & \textbf{Expected} \\
\midrule
intake/plan      & request becomes story + checklist + verification intent & \textsc{pass} \\
executor         & executor follows typed planned steps within scope        & \textsc{pass} \\
repair-loop      & a failure is repaired within the retry budget            & \textsc{pass} \\
review-loop      & a review finding is handled via a bounded loop           & \textsc{pass} \\
evidence-gate    & missing evidence blocks; fresh evidence unblocks         & \textsc{block} \\
block-escalate   & an unsafe/ambiguous task does \emph{not} advance as done & \textsc{block} \\
budget-stop      & time/cost/iteration caps stop the loop                   & \textsc{block} \\
human-handoff    & a human decision is requested when required              & \textsc{block} \\
multi-host-review& independent host verdicts are required (local-only here) & \textsc{pass} \\
no-false-done    & missing proof cannot become done                         & \textsc{block} \\
\bottomrule
\end{tabular}
\end{table}

\paragraph{Large-ledger no-false-\textsc{done} stress suite.} The ten-scenario contract above
checks the loop's lifecycle surface. A second Tier-A stress suite checks a different failure
mode: large user intent ledgers. It instantiates ten synthetic parent ledgers, each with exactly
15 required milestone rows, for a 150-row local matrix. The rows are intentionally
\emph{milestone-ledger rows}, not 150 real implementation stories: the purpose is to prove that
the gate can distinguish coverage, freshness, deferral, assignment, and claim-boundary states
before allowing a \textsc{done} summary.

\begin{table}[t]
\centering\scriptsize
\caption{UDM large-ledger no-false-\textsc{done} baseline (Tier A). Each group contains
15 required milestone-ledger rows; the aggregate smoke asserts exactly $10\times15=150$
required rows. The suite is executable local evidence, not an empirical superiority claim.}
\label{tab:udm-ledger-stress}
\begin{tabularx}{\textwidth}{@{}p{0.75cm}Y Y@{}}
\toprule
\textbf{Group} & \textbf{Constructed ledger condition} & \textbf{Capability proved} \\
\midrule
G01 & All 15 rows closed with fresh machine evidence &
Positive control: the ledger can reach full completion when every required row has admissible
evidence. \\
G02 & 14 rows closed and one required row missing &
One missing milestone cannot be hidden by an otherwise large pass set; 14/15 is not
\textsc{done}. \\
G03 & 15 rows carry prose completion text but no machine evidence &
Documentation or narrative closure cannot satisfy a delivery claim. \\
G04 & 15 rows contain stale evidence bindings &
Evidence freshness is enforced; stale material/story/head bindings block completion. \\
G05 & 12 rows closed and 3 rows human-approved for deferral &
Deferral remains visible and open; approved deferral is not silently counted as implemented
runtime delivery. \\
G06 & 15 migration/sample rows under legacy or candidate semantics &
Historical samples and migration scaffolds cannot become current roadmap completion evidence. \\
G07 & 15 local rows closed but the summary attempts stronger wording &
Local deterministic proof cannot be over-claimed as production, Authority, fullAuthority, or
independent-host completion. \\
G08 & Duplicate or malformed row identifiers are present &
Malformed or duplicate rows cannot cover distinct user commitments; row identity is checked. \\
G09 & Every row is assigned to a child story but lacks delivery evidence &
Assignment is not delivery; a scheduled child story cannot close the parent ledger by itself. \\
G10 & 15 rows start stale, then are refreshed to current evidence &
The loop can recover safely: completion remains blocked before refresh and unlocks only after
fresh evidence is bound. \\
\bottomrule
\end{tabularx}
\end{table}

\subsection{Reflection-loop ablation (powered: 9{,}240 cells)}\label{sec:loopB}

\begin{table}[H]
\centering\small
\caption*{\textbf{Evaluation setup at a glance for the powered ablation.}}
\begin{tabularx}{\textwidth}{@{}p{3.7cm}X@{}}
\toprule
\textbf{Item} & \textbf{Setting} \\
\midrule
Model & Sonnet provider model label; exact dated provider snapshot not recorded in the
experiment records. \\
Arms & A1 prompt-only; A2 naive-retry; A2$'$ compute-budgeted naive; A3 review-only;
A4 Proof-or-Stop loop. \\
Tasks & 24 stratified tasks. \\
Scenarios & Null plus B1--B15 injected-failure scenarios. \\
Repeats & $k=5$. \\
Total cells & 9{,}240 applicable unique cells: A1 covers 13/16 scenarios; A2/A2$'$/A3/A4
cover 16/16; 0 invalid records. \\
Primary endpoint & Not-amplified rate. \\
Secondary endpoints & Completion, model/tool cost, token count, and wall time. \\
Statistical readout & Wilson 95\% CIs~\cite{wilson1927}, seeded cluster
bootstrap~\cite{efron1979}, and exploratory Benjamini--Hochberg
FDR~\cite{benjamini1995} over per-scenario tests. \\
Caveats & One model, no external benchmark result, and A2$'$ is compute-budgeted rather than
perfectly equal-spend per run. \\
\bottomrule
\end{tabularx}
\end{table}

To evaluate whether stronger control improves outcomes relative to weaker loops, we
pre-register a \textbf{five-arm} ablation~\cite{meyes2019ablation}. All arms used the same
provider model family/model label, tool surface, tasks, and randomized run window; the exact
dated provider snapshot was not recorded, which is a reproducibility limitation. Thus the
\emph{control logic} is the intended difference:
(A1) \emph{prompt-only} (one pass, no loop);
(A2) \emph{naive-retry} (blind retry on failure, $R{=}3$, no gate);
(A2$'$) \emph{compute-budgeted naive} --- identical to A2 but bounded by a \emph{token $+$
wall-clock spend budget} equal to A4's per-task median (measured in pilot; $\pm$20\% band, token
binds first, hard-truncated), so it is \emph{compute-budgeted to A4's pilot median} rather than
structured;
(A3) \emph{review-only} (A2 plus exactly one review pass using A4's reviewer, not iterated); and
(A4) the \emph{Proof-or-Stop reflection loop} (plan$\to$execute$\to$review$\to$bounded
reflection$\to$evidence gates$\to$done).

\begin{figure}[t]
\centering
\includegraphics[width=0.92\textwidth]{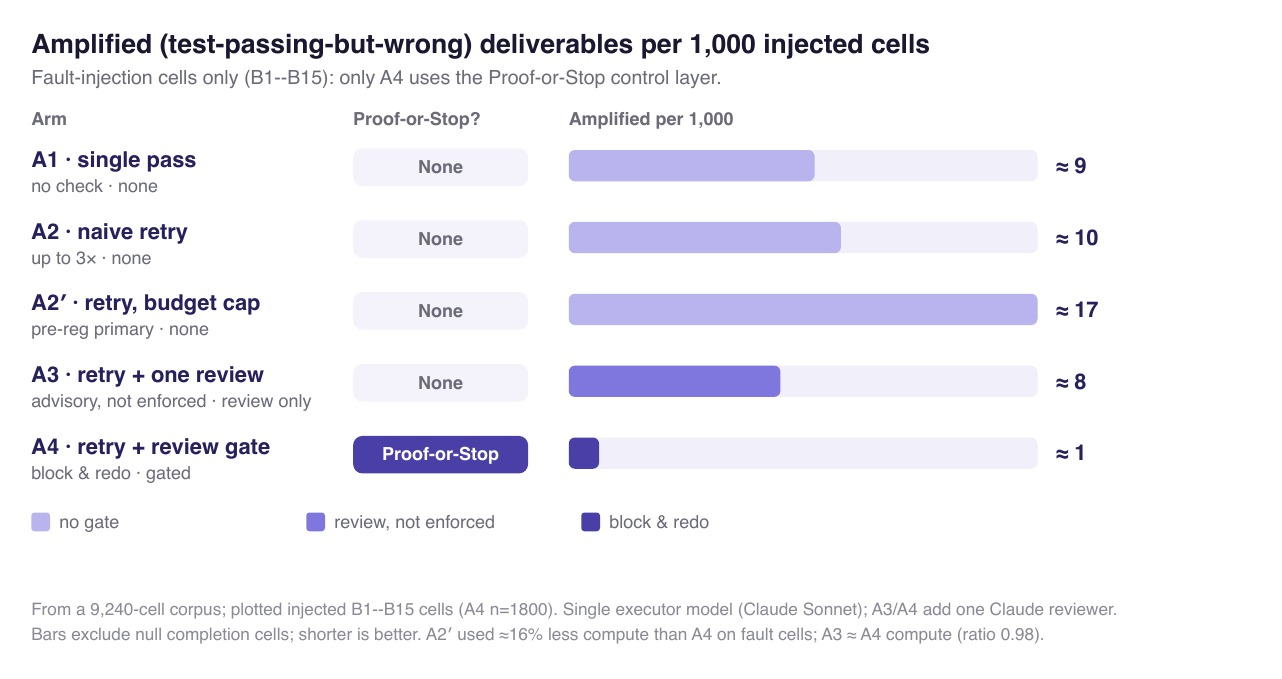}
\caption{Amplification-rate ladder for the powered ablation (computed over injected B1--B15
cells only; lower is better). The middle column marks whether the arm uses the Proof-or-Stop
control layer: A1--A3 are \emph{None}, while A4 is \emph{Proof-or-Stop}. Bars report
visible-test-passing but hidden-failing artifacts per
1{,}000 injected cells. The pre-registered primary contrast is A4 Proof-or-Stop loop versus A2$'$
compute-budgeted naive: A4 amplifies $2/1800$ cases (${\approx}1.1$/1{,}000), while A2$'$
amplifies $31/1800$ cases (${\approx}17.2$/1{,}000), with H1 not-amplified $+1.6$pp
[0.8,\,2.5], CI excluding 0. A1 prompt-only is a reference arm
($13/1440$, ${\approx}9.0$/1{,}000).
The arms are categorical mechanisms, so we plot a bar ladder rather than a continuous curve.}
\label{fig:powered-ladder}
\end{figure}

\paragraph{What counts as wrong.}
Each programming task has two scoring surfaces. The \emph{visible acceptance test} is available to
the agent and is the test a weak loop can learn to satisfy. A hidden ground-truth oracle, kept out
of the agent prompt and used only by the harness, decides whether the accepted artifact is actually
correct. For the non-null scenarios B1--B15, the harness injects failures designed to expose this
gap: an artifact may pass the visible acceptance check while still failing the hidden oracle. We
score such a shipped artifact as \textsc{amplified}. Thus the ``wrong'' outcome in
Fig.~\ref{fig:powered-ladder} is not a subjective review label; it is a machine-read event:
visible acceptance passes, the hidden correctness oracle fails, and the arm nevertheless ships or
propagates the artifact. The complement, \emph{not-amplified}, means the arm either repairs the
artifact or refuses to advance it.

\textbf{A2$'$ is the pre-registered budget-capped naive control}: it reduces, but does not eliminate,
the raw-compute confound, so the headline comparison \textbf{A4 vs A2$'$} cannot be read as a clean
equal-spend causal estimate. Cost (model calls $+$ tokens $+$ wall-clock) is a \emph{primary co-metric}, reported
beside completion so more compute is never hidden. Each arm runs over $24$ stratified tasks
$\times\,5$ repeats; outcomes are read \emph{objectively} from git history, the process-event
log, and an independently-authored, known-good-validated acceptance script --- never from agent
narration. For each rate $p$ with $n$ trials we report a Wilson 95\% interval~\cite{wilson1927}
\begin{equation}
\widehat{p}_\pm \;=\; \frac{\widehat{p} + \frac{z^2}{2n} \pm z\sqrt{\frac{\widehat{p}(1-\widehat{p})}{n} + \frac{z^2}{4n^2}}}{1 + \frac{z^2}{n}},\qquad z=1.96,
\label{eq:wilson}
\end{equation}
and test arm differences with a two-proportion test (Fisher's exact~\cite{fisher1922}
per scenario given small
per-cell $k$ and large expected effects), reporting effect size with CI rather than $p$ alone.
The pre-registered hypotheses are \textbf{H1} (A4 not-amplified rate $>$ A2$'$, compute-budgeted control),
\textbf{H2} (A4 completion $\ge$ A2$'$ \emph{and} A4 regression $<$ A2$'$), and \textbf{H3} (A4
$>$ A3 --- does the \emph{bounded loop} add value beyond a \emph{single} review?).
\textbf{Status: powered run COMPLETE (9{,}240 applicable unique cells).} The design starts from
5 arms $\times$ 24 stratified tasks $\times$ 16 scenarios $\times\,k{=}5$; per-arm applicability
gives A1 13 scenarios (1{,}560 cells) and A2/A2$'$/A3/A4 all 16 scenarios
(7{,}680 cells), for 9{,}240 scored cells total (Sonnet; 0 invalid records). The pre-registered
primary contrast \textbf{A4 vs A2$'$} (compute-budgeted naive control, seeded cluster bootstrap, $B{=}2000$) gives
\textbf{H1 (\S\ref{sec:recovery} not-amplified) $=+1.6$pp, 95\% CI $[0.8,2.5]$} --- the CI
excludes zero: the evidence-gated loop amplifies \emph{less} than the compute-budgeted naive
loop, and does so while spending \emph{more} (A4 $\approx\!1.2\times$ A2$'$ tokens), a conservative
direction. \textbf{H2 (completion) $=+3.3$pp $[0.0,10.0]$} is marginal (null-cell completion is
near-ceiling for every arm). Full per-arm rates are in Table~\ref{tab:powered-ablation}; the
clean-task pilot (mechanism, no injection) is reported below.
The deterministic execution+analysis \emph{harness} is built, offline-tested, and locally
3$\times$2 host-verdict reviewed. The harness (in the recovery-runner, merged to Proof-or-Stop main) is the full pipeline
\emph{enumerate $\to$ run (pluggable engine) $\to$ adjudicate $\to$ score $\to$ analyze}: a
cell-matrix driver honouring the \S\ref{app:protocol} applicability table, the \S4.2 total-order
adjudicator, harness-owned hash-checked acceptance $+$ hidden-regression $+$ diff-hygiene scoring,
and an analysis stage that computes \emph{matched} common-cell rates with the Wilson interval of
Eq.~\eqref{eq:wilson}, an A4-vs-A2$'$ seeded cluster bootstrap as the primary readout, one
Benjamini--Hochberg FDR over the exploratory per-scenario tests, the cost co-metric,
and an emitted \code{analysis.glmm.R} sensitivity script (two disjoint H1/H2 \code{lme4::glmer} models~\cite{bates2015lme4} over the matched
corpus); the harness's lifecycle-readiness evidence (the live-engine adapter B-8d44, self-checks,
and a local review quorum) is in Appendix~\ref{app:protocol}. A
\emph{matched-cell} invariant --- every per-arm estimate scores both arms over the same
both-arms-have-data cell set, so no headline number can be inflated by a mismatched denominator
(\S\ref{sec:selfapp}) --- closes the \S8 denominator-bias threat in code. The full protocol
(arms, stratified corpus, per-arm injection applicability, readout, validity filters) is in
Appendix~\ref{app:protocol}.

\begin{table}[H]
\centering\small
\caption{\textbf{Powered ablation result} (9{,}240 applicable unique cells from a
5-arm $\times$ 24-task $\times$ 16-scenario $\times\,k{=}5$ design after per-arm applicability:
A1 covers 13 scenarios; A2/A2$'$/A3/A4 cover all 16; Sonnet; 0 invalid records). Completion is over
matched null (no-injection) cells; amplified counts are visible-pass/hidden-fail outcomes over
matched injected B1--B15 cells (lower is better). The pre-registered H1 statistic is the
not-amplified endpoint; for the primary A4-vs-A2$'$ cells it is exactly the complement of
amplified outcomes. Both use the \S8 matched-denominator invariant. Cost is the per-cell mean.
Rows are ordered by loop fidelity; A2$'$ is the pre-registered primary control but not a clean
causal estimate of budgeting, since realized spend matching is coarse and its rare-event result is
weaker than raw A2 in this run. Fig.~\ref{fig:powered-ladder} shows the same result on the rarer
amplification scale.}
\label{tab:powered-ablation}
\begin{tabular}{@{}lcccr@{}}
\toprule
\textbf{Arm ($\uparrow$ fidelity)} & \textbf{Completion (null)} & \textbf{Amplified (B1--B15, $\downarrow$)} & \textbf{Mean tokens} & \textbf{Mean wall} \\
\midrule
A1 prompt-only        & 93.3\% [87.4,96.6] & 13/1440 (0.90\%) & 168{,}622 & 51.7\,s \\
A2 naive-retry        & 100\% [96.9,100]   & 18/1800 (1.00\%) & 169{,}202 & 51.3\,s \\
A2$'$ compute-budgeted & 96.7\% [91.7,98.7] & 31/1800 (1.72\%) & 170{,}545 & 54.8\,s \\
A3 review-only        & 100\% [96.9,100]   & 14/1800 (0.78\%) & 200{,}510 & 80.3\,s \\
A4 Proof-or-Stop loop    & 100\% [96.9,100]   & \textbf{2/1800 (0.11\%)} & 204{,}553 & 81.2\,s \\
\bottomrule
\end{tabular}
\vspace{0.2em}
\begin{flushleft}\scriptsize
\textbf{Primary A4$-$A2$'$} (compute-budgeted naive control, seeded cluster bootstrap, $B{=}2000$): \textbf{H1}
not-amplified $+1.6$pp [0.8,\,2.5] (\emph{CI excludes 0}); \textbf{H2} completion
$+3.3$pp [0.0,\,10.0] (marginal). H2's second prong (regression $<$ A2$'$) is a \emph{no-injection}
secondary and is $0/0$ for every arm on the null cells, so it is not separately estimable here;
the injection-bearing hidden-regression failures coincide with the amplification counts above
(A4 $=2$, A2$'$ $=31$ over the B1--B15 injected cells). Because amplification is a rare event,
the amplified count/rate is the interpretable display; the pre-registered not-amplified endpoint
is its complement for this primary contrast. \emph{Not-amplified $=$ recovered $+$ safe-stop}; safe-stop $=0$ for every arm here
(the full A4 loop repairs rather than merely stopping), so not-amplified $=$ recovered for A2/A2$'$/A3/A4.
A1 has 16 injected single-shot acceptance-fail edge cells outside the
recovered/amplified/safe-stop buckets in this B1--B15 readout; across all A1 cells the total is
24 (16 injected, 8 null/no-injection), and A1 is only a reference arm. A2$'$ is compute-budgeted to A4's pilot median, but actual per-run spend matching is coarse; we
therefore report realized token and wall-clock costs. A4 wins H1 while spending $\approx\!1.2\times$
A2$'$ tokens (conservative); only $207/1920$ all-scenario A2$'$ cells land inside the $\pm20\%$
band (single-round token granularity), so the contrast rests on A4's higher \emph{mean} spend,
not per-run equality. Amplification is rare
and spread thinly: A4 $=2/1800$ vs A2$'$ $=31/1800$ events across B1--B15; exploratory per-scenario
Benjamini--Hochberg $0/15$ rejected (small per-cell $k$; no per-scenario contrast reached
significance). \S4.2 unsafe detectors (forbiddenPath/secretScan/perms/forbiddenCreate/destructiveCmd)
each fired $0$ --- amplification here is shipping a visible-pass/hidden-fail artifact, not an unsafe
operation.
\end{flushleft}
\end{table}

\paragraph{Cost--reliability trade-off.} Table~\ref{tab:powered-ablation} shows the main
operational trade-off: stronger control reduces visible-pass/hidden-fail amplification, but it
costs more tokens and time. A4 amplifies $2/1800$ injected cases versus $31/1800$ for the
pre-registered budget-capped A2$'$ control (equivalently, H1 not-amplified $+1.6$pp
[0.8,2.5]) while using 204{,}553 mean tokens and 81.2\,s per cell, compared with 170{,}545
tokens and 54.8\,s for A2$'$. This supports the narrower
claim that Proof-or-Stop is a reliability-oriented control mechanism, not a free speed-up or a clean
equal-spend result.
Table~\ref{tab:powered-ablation} reports the pre-registered not-amplified endpoint and cost
co-metrics; Fig.~\ref{fig:powered-ladder} plots the same injected B1--B15 evidence on the
amplified-outcomes scale, making the rare-event A4-vs-A2$'$ contrast explicit.

\paragraph{What the ablation proves.}
The powered result is a mechanism claim, not a general theorem that agents become correct. First,
the enforced gate helps against the pre-registered budget-capped naive control A2$'$: A4 reduces
amplification from $31/1800$ to $2/1800$ injected cells, yielding H1 not-amplified $+1.6$pp with a
95\% cluster-bootstrap CI $[0.8,2.5]$ that excludes zero. This contrast is pre-registered but not a
clean causal estimate of budgeting: A2$'$ realized-spend matching is coarse, A4 spends more on
average, and A2$'$ is weaker than raw A2 in this rare-event run. Second, the A3--A4 pair is the
cleanest exploratory isolation of enforcement: A3 spends nearly the same tokens as A4
(A3/A4 token ratio $0.98$) and uses the same one-review signal, but A3 treats the review as
advisory and amplifies $14/1800$ injected cells; A4 enforces the review gate and amplifies
$2/1800$. This enforcement-isolation contrast is less concentrated than the A2$'$ comparison:
excluding \code{du-duration}, A3 still amplifies $9/1725$ cells while A4 amplifies $0/1725$
cells. Third, this powered readout measures recovery/not-amplification, not terminal stop
activation: safe-stop and the unsafe-action detectors are zero in these records, so stop behavior is
supported by the separate engine-contract and recovery-pilot evidence. The supported conclusion is
therefore narrow: for this pre-registered coding corpus, Proof-or-Stop-style enforcement reduces
visible-pass/hidden-fail error amplification relative to weaker control logic, with the strongest
primary-control contrast occurring where the visible-pass/hidden-fail trap is active.

\paragraph{What the powered ablation instantiates.}
The powered ablation instantiates the \emph{control-policy} contrast over a deterministic experiment
harness. It does not instantiate the full story-level materialHash/receipt-gated lifecycle for each
cell. The story-level Proof-or-Stop engine tests separately validate materialHash freshness,
\textsf{commandSetHash} binding, full-test receipts, review verdict admissibility, and the
\textsc{done} gate; the powered ablation validates whether enforcing the review-and-redo gate
reduces visible-pass/hidden-fail amplification under a fixed task/scenario matrix. These are
complementary evidence objects rather than the same artifact. A future gate-grade replay should
package selected powered cells as story-level \mh/\textsf{commandSetHash}/receipt evidence.

\paragraph{Supplemental paired execution-status comparison.}
As a descriptive companion to the powered control-policy result, we joined a separate executed
Proof-or-Stop gated run and a no-review control run by task, scenario, and repeat over the same
1{,}152 cells. The no-review control reached terminal completion on $1{,}143/1{,}152$ cells, while
Proof-or-Stop admitted $1{,}042/1{,}152$ cells after completion or recovery and safe-stopped
$110/1{,}152$ cells. Among paired cells, 106 no-review completions were not admitted by
Proof-or-Stop (Table~\ref{tab:cell03-cell06-paired}). Because these safe-stops were not
adjudicated against hidden ground truth, we do not claim that the 106 no-review artifacts were
incorrect. The supported observation is narrower: terminal completion and admissible delivery are
distinct lifecycle states, consistent with the agent-as-claim framing. This is a separate
execution-status matrix from the multi-model-ablation program (Phase 1; protocol incomplete) and
does not extend the pre-registered \S\ref{sec:loopB} powered ablation result. Full precision and
native outcome details are in Appendix~\ref{app:cell03-cell06}.

The difference from the safe-stop $=0$ powered-ablation readout in Table~\ref{tab:powered-ablation}
is protocol-driven rather than contradictory: in the powered matrix A4 repaired the injected
visible-pass/hidden-fail cases within that harness, whereas this execution-status matrix records a
safe-stop whenever the gated run's evidence path does not admit delivery.

\paragraph{Supplemental token-usage readout.}
The same paired matrix also provides a descriptive token-usage comparison. Under input+output token
semantics, without double-counting cached-input or reasoning-output subfields, the gated run used
221{,}068{,}475 input+output tokens over 1{,}152 matched final rows, while the no-review control used
58{,}173{,}502 input+output tokens over the same 1{,}152 matched formal rows. This gives a
3.80$\times$ input+output token-usage ratio and an incremental 162{,}894{,}973 input+output tokens.
Both sides are labeled as OpenAI/GPT-family \code{gpt-5.5} runs in the underlying artifacts; if a
future validation showed a model-family mismatch, the ratio should be replaced by side-by-side
provider-reported counts. This is a bundled condition comparison, not an isolated estimate of review
overhead, not a dollar-cost estimate, and not a cost-benefit proof. Cell06 reports cached input
separately (46{,}874{,}624 of 57{,}199{,}341 input tokens, 81.95\%), while Cell03 does not expose a
matching cached-input breakdown, so the ratio is not necessarily cost-proportional under provider
billing. This supplemental 3.80$\times$ token-usage ratio is not directly comparable to the
approximately 1.2$\times$ token ratio in Table~\ref{tab:powered-ablation}: the powered ablation uses
a different pre-registered matrix and a compute-budgeted A2$'$ control, while this paired readout
compares a bundled gated run with a no-review control in a separate execution-status matrix.

\begin{table}[H]
\centering\small
\caption{Supplemental paired completion-vs-delivery matrix over the same 1{,}152 cells, joined by
task, scenario, and repeat. Rows show the terminal status of the no-review control; columns show
the Proof-or-Stop delivery decision. Counts are descriptive and not hidden-oracle adjudicated.
Appendix~\ref{app:cell03-cell06} reports full precision.}
\label{tab:cell03-cell06-paired}
\begin{tabular}{@{}lrr@{}}
\toprule
\textbf{No-review control} & \textbf{Proof-or-Stop admitted} & \textbf{Proof-or-Stop safe-stopped} \\
\midrule
Completed & 1{,}037 & 106 \\
Failed & 5 & 4 \\
\bottomrule
\end{tabular}
\end{table}

\paragraph{Readout completeness and deviations.} The headline A4-vs-A2$'$ estimate is the
in-process seeded cluster bootstrap. The pre-registered GLMM script is \emph{emitted} as
\code{analysis.glmm.R} (two disjoint H1/H2 \code{lme4::glmer} models~\cite{bates2015lme4} over the
matched corpus) for external fitting, but is not used as an inferential claim in this paper; the
primary inferential claim is the seeded cluster bootstrap. \textbf{H3} (A4 vs A3,
\emph{near-compute but not strict per-run matched}: A4 not-amplified
$99.9\%$ vs A3 $99.2\%$ at A4 $\approx$ A3 spend) is exploratory. The secondary descriptors
(uncontrolled-retries, evidence-completeness, human-intervention) and the per-category breakdown
are not separately tabulated: under the single pre-registered FDR no per-scenario contrast reached
the FDR threshold, and the safe-stop / event-based detectors are structurally $0$ in these records.
\emph{Data hygiene:} 77 restart duplicates (identical outcomes, from the multi-day run) were
de-duplicated by cell key (arm$|$task$|$scenario$|$repeat) to reach the $9{,}240$ unique cells, and
24 A1 (single-shot) acceptance-fail edge cells (16 injected, 8 null/no-injection) fall outside the
recovered/amplified/safe-stop classification and sit in the non-primary A1 reference arm. A4's two amplifications both fall on a
single task (\code{du-duration}); every other task is $100\%$ not-amplified for A4.
Table~\ref{tab:du-duration-sensitivity} shows the concentration explicitly. The primary effect
should therefore be read as a matrix-level aggregate, not per-task dominance: Proof-or-Stop buys
little on easy cells that almost never amplify, and most visibly helps on the task where the
visible-pass/hidden-fail trap is active. The released
records carry \code{modelId}=\code{sonnet} (the exact dated provider snapshot was not recorded in
the experiment records) and
leave \code{headHash}/\code{materialHash} null --- the powered harness does not material-hash-gate
experiment cells (distinct from the story-level evidence gate that does). Thus the scoring
artifacts are mechanically reproducible from released records, but model-output replay is not
provider-snapshot reproducible and these records are not lifecycle-gate-grade evidence under
Eq.~\eqref{eq:adm}. Raw records, the tidy CSV, \code{analysis.glmm.R}, a per-scenario CSV, and
scenario/task/schema dictionaries are released in
\code{experiments/powered-ablation/}.

\begin{table}[H]
\centering\small
\caption{Task-concentration sensitivity for the primary A4-vs-A2$'$ amplified-outcome contrast.
Counts are over injected B1--B15 cells from the released \code{analysis.tidy.csv}; lower is better.}
\label{tab:du-duration-sensitivity}
\begin{tabularx}{\textwidth}{@{}p{3.0cm}p{2.6cm}p{3.0cm}Y@{}}
\toprule
\textbf{Scope} & \textbf{A4 Proof-or-Stop} & \textbf{A2$'$ compute-budgeted naive} & \textbf{Reading} \\
\midrule
All 24 tasks & $2/1800$ amplified & $31/1800$ amplified &
A4 releases far fewer visible-pass/hidden-fail artifacts in the full powered matrix. \\
\code{du-duration} only & $2/75$ amplified & $29/75$ amplified &
Most of the aggregate separation comes from the highest-risk task, where the gate prevents many
wrong artifacts that naive retry ships. \\
Excluding \code{du-duration} & $0/1725$ amplified & $2/1725$ amplified &
The remaining tasks are near-ceiling for both arms; the effect is small because there are few
errors left to prevent. \\
\bottomrule
\end{tabularx}
\end{table}

\paragraph{Pilot (clean-task ablation, $n{=}9$/arm).} As a first measurement we ran all four
arms on three clean, well-specified tasks (\code{parse-range}, \code{median}, \code{text-stats}),
$k{=}3$, \emph{no injection} --- the deliberate complement of the \S\ref{sec:recovery} fault
case (Table~\ref{tab:ablation-pilot}). All four arms completed correctly $9/9$ (hidden
ground-truth pass) in a single attempt: naive-retry never retried and bounded reflection never
triggered, because there was no failure to recover from (A1$\approx$A2, A3$\approx$A4). The only
separation is cost and latency --- the independent review gate roughly doubles both (A4/A1 $=
2.07\times$ cost, $2.51\times$ wall). This is the honest boundary of the loop: \emph{on easy,
well-specified work the loop is pure overhead with no completion benefit}; its value appears
precisely under the green-but-wrong faults of \S\ref{sec:recovery}, where the bare loop amplified
$15/15$ and the gated loop shipped $0/15$ wrong results. The loop functions as a
risk-mitigation mechanism: it adds overhead on clean tasks, but becomes valuable when the visible
test is green and the artifact is wrong. \emph{Caveat:}
$n{=}9$/arm, three small tasks, one model; with $9/9$ the completion CI is wide
($[0.70,1.0]$), so this bounds completion at ``no observed failures,'' not a powered equality ---
the cost/latency ladder is the robust signal; the powered fault-bearing study (H1--H3) is now
complete (Table~\ref{tab:powered-ablation}: H1 $+1.6$pp [0.8,2.5], CI excluding 0).

\begin{table}[H]
\centering\small
\caption{Clean-task ablation pilot ($n{=}9$/arm; 3 tasks $\times\,k{=}3$; no injection; Sonnet;
completion = hidden ground-truth pass). All arms complete; loop fidelity adds cost/latency with
no completion benefit --- the loop's value is the fault case (\S\ref{sec:recovery}), not easy
clean work.}
\label{tab:ablation-pilot}
\begin{tabular}{@{}lccrr@{}}
\toprule
\textbf{Arm} & \textbf{$n$} & \textbf{Completion (95\% CI)} & \textbf{Mean cost} & \textbf{Mean wall} \\
\midrule
A1 prompt-only   & 9 & $9/9$ \,[0.70, 1.0] & \$0.090 & 16.6\,s \\
A2 naive-retry   & 9 & $9/9$ \,[0.70, 1.0] & \$0.094 & 19.1\,s \\
A3 review-only   & 9 & $9/9$ \,[0.70, 1.0] & \$0.185 & 44.5\,s \\
A4 Proof-or-Stop loop  & 9 & $9/9$ \,[0.70, 1.0] & \$0.186 & 41.7\,s \\
\bottomrule
\end{tabular}
\end{table}

\section{Recovery under Injected Failure}\label{sec:recovery}

To evaluate behavior under injected failures, we separate the Tier-A contract result from
comparative recovery measurements. The engine contract demonstrates that the loop can stop,
escalate, and repair; the empirical question is the rate at which it does so under realistic
faults. We pre-register fifteen deliberate failure injections (Table~\ref{tab:recovery}), each
run under \{naive-retry loop, Proof-or-Stop loop\} (optionally a review-only arm) with $k\ge 3$
repeats. Each injection has an \emph{objective} readout derived from the final diff, the
process-event log, or the gate decisions --- e.g.\ ``does the final diff contain the
out-of-scope file?'', ``did the loop stop after $N$ no-progress iterations or spin forever?''.
The deliverable is, per scenario, ``Proof-or-Stop loop contains/recovers $X\%$ vs naive $Y\%$, $\Delta$
with 95\% CI'' (Eq.~\eqref{eq:wilson}). We deliberately drop injections that merely re-assert
the Tier-A contract checks, keeping the net-new empirical scenarios.
\textbf{Status: powered run COMPLETE (9{,}240 cells).} Over the full matrix, the evidence-gated
loop A4 is not-amplified $99.9\%$ [99.6,100] vs the compute-budgeted naive loop A2$'$ $98.3\%$
[97.6,98.8] (primary \textbf{H1 $=+1.6$pp [0.8,2.5]}, CI excluding 0;
Table~\ref{tab:powered-ablation}). Amplification is rare and spread thinly across scenarios
(A4 $=2$, A2$'$ $=31$ events over B1--B15), so the powered run contributes statistical power over
the full matrix. It complements the discrimination-cell pilot below, which isolates the
mechanism with a large effect on selected green-but-wrong cells. An exploratory single-agent
probe over all fifteen injections (Table~\ref{tab:recovery}, last column) illustrates the
targeted failure mode: the bare agent amplified on 5/15 (e.g.\ garbage-fix, delete-critical,
destructive-migration), recovered on 6, and safe-stopped on 4 ($n{=}1$ per scenario, no arm
comparison---illustrative only, not the powered result, which is now in
Table~\ref{tab:powered-ablation}).

The pilot separates loop-fidelity levels before the full lifecycle harness: A2$'$ is a
compute-budgeted bare loop, A4-C is a single independent review gate, and A4b-B is a B-fidelity
proxy of the A4 loop used to isolate the block$\to$revise$\to$re-review mechanism.

For the pilot, let $A_j$ denote visible-acceptance pass, $G_j$ hidden ground-truth pass, and
$S_j$ whether the arm ships the final artifact (for A2$'$, the bare loop ships by construction;
for gated arms, $S_j$ is the final independent-review \textsc{ship} verdict). We score each run by
\begin{equation}
\mathrm{Out}(j)=
\begin{cases}
\textsc{completed/recovered}, & S_j \land G_j,\\
\textsc{amplified},           & S_j \land \neg G_j,\\
\textsc{safe\mbox{-}stop},    & \neg S_j \land \neg G_j,\\
\textsc{false\mbox{-}stop},   & \neg S_j \land G_j.
\end{cases}
\label{eq:recovery-outcome}
\end{equation}
The non-definitional discrimination cells are exactly $A_j \land \neg G_j$: the visible test is
green, but the artifact is wrong under hidden ground truth, so only an independent gate can
prevent amplification.

\begin{table}[H]
\centering\scriptsize
\setlength{\tabcolsep}{3pt}
\caption{Main recovery result under ``passes-the-visible-test-but-wrong'' injection. Pilot:
3 tasks (\code{parse-range}, \code{median}, \code{text-stats}) and 5 task-injection cells,
$n{=}3$ each, for 15 wrong-injection runs per executed arm; Sonnet, B-fidelity proxy,
independently verified by V3/V4 audits. \emph{pre-reg.} marks reference arms from the protocol
that were not executed in this pilot table; \textsc{n.r.} means the no-injection column was not run
in this pilot.}
\label{tab:recovery-main-result}
\begin{tabularx}{\textwidth}{@{}>{\raggedright\arraybackslash}p{3.15cm}
>{\centering\arraybackslash}X
>{\centering\arraybackslash}X
>{\centering\arraybackslash}X
>{\centering\arraybackslash}X
>{\centering\arraybackslash}X
>{\centering\arraybackslash}X
>{\centering\arraybackslash}X@{}}
\toprule
\textbf{Arm ($\uparrow$ loop fidelity)} &
\textbf{Amplified} & \textbf{Safe-stop} & \textbf{Recovered} &
\textbf{Completion} & \textbf{False-DONE} & \textbf{Cost/run} & \textbf{Wall/run} \\
\midrule
A1 prompt-only & \emph{pre-reg.} & \emph{pre-reg.} & \emph{pre-reg.} &
\emph{pre-reg.} & \emph{pre-reg.} & \textsc{n.r.} & \textsc{n.r.} \\
A2 naive-retry & \emph{pre-reg.} & \emph{pre-reg.} & \emph{pre-reg.} &
\emph{pre-reg.} & \emph{pre-reg.} & \textsc{n.r.} & \textsc{n.r.} \\
A2$'$ compute-budgeted bare & \textbf{15/15} & 0 & 0 &
\textsc{n.r.} & \textsc{n.r.} & \$0.094 & 18.3s \\
A3 review-only & \emph{pre-reg.} & \emph{pre-reg.} & \emph{pre-reg.} &
\emph{pre-reg.} & \emph{pre-reg.} & \textsc{n.r.} & \textsc{n.r.} \\
A4-C one review gate & 0 & \textbf{15/15} & 0 &
\textsc{n.r.} & \textsc{n.r.} & \$0.173 & 34.8s \\
A4b-B bounded reflection & 0 & 0 & \textbf{15/15} &
\textsc{n.r.} & \textsc{n.r.} & \$0.386 & 89.0s \\
\bottomrule
\end{tabularx}
\vspace{0.2em}
\begin{flushleft}\scriptsize
Mean work/run for executed arms: A2$'$ = 1.00 attempts; A4-C = 1.00 attempts plus one review
gate; A4b-B = 2.067 attempts and 1.067 review rounds.
\end{flushleft}
\end{table}
In this pilot, the loop is not free; under green-but-wrong faults, added evidence and review
fidelity act as a risk-mitigation mechanism, moving outcomes from amplification to safe-stop to
bounded repair.

\begin{figure}[t]
\centering
\includegraphics[width=0.86\textwidth]{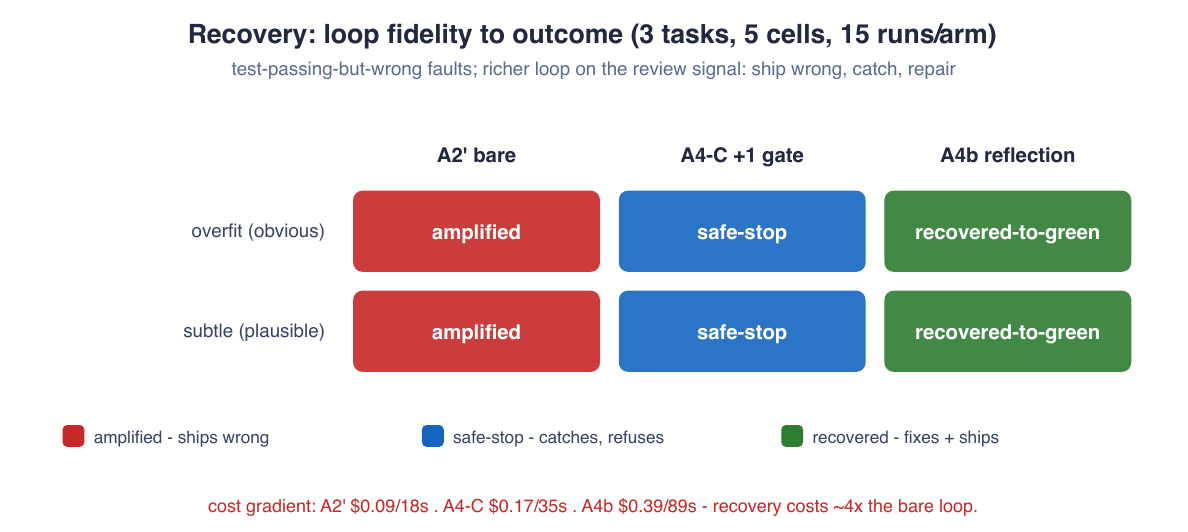}
\caption{Verified loop-fidelity gradient on visible-test-passing wrong artifacts across three
tasks and five task-injection cells (15 runs per arm). A2$'$ has only the visible test and ships
the wrong artifact; A4-C adds one independent gate and refuses to ship; A4b-B feeds the review
finding back through bounded reflection and repairs before shipping.}
\label{fig:recovery-fidelity}
\end{figure}

\paragraph{Pilot (verified proof-of-mechanism, $n{=}3$ per task-cell, three tasks).} A pilot
confirms the measurement pipeline runs end-to-end and exercises the \emph{thesis-shaped}
differential at increasing difficulty (Table~\ref{tab:recovery-main-result};
Fig.~\ref{fig:recovery-fidelity}).
Early delete/corrupt cells verified the classifier and runner, but their differential is partly
definitional: A4 safe-stops when acceptance is not met, while A2$'$ ships whatever it has. The
non-definitional signal comes from the $A_j\land\neg G_j$ cells in
Eq.~\eqref{eq:recovery-outcome}. Across \code{parse-range}, \code{median}, and \code{text-stats},
the injected artifacts pass visible acceptance while failing hidden ground truth. A2$'$ ships these
wrong artifacts 15/15 (\textbf{amplified}); A4-C catches them and refuses to ship 15/15
(\textbf{safe-stop}); A4b-B consumes the same review signal through bounded reflection and repairs
them 15/15 (\textbf{recovered-to-green}). The subtle \code{parse-range} defect is a complete
plausible implementation whose falsy guard \code{if(!n)} wrongly rejects \code{0}; the
\code{median} subtle defect returns the lower middle element rather than the average for
even-length arrays; the \code{text-stats} subtle defect is buried in a 63-line multi-helper
artifact. Captured reviewer reasoning names these defects and their fixes.

Separate V3/V4 data/code audits verified that A4b-B recovery is not a classifier artifact: all 15
A4b-B runs have \code{groundTruthPass=true}, final reviewer \textsc{ship}, and
\code{reviewRounds>=1}; in the runner, a review round increments only after a reviewer
\textsc{block}, while \textsc{recovered-to-green} requires a final shipped artifact that passes
hidden ground truth. Thus the key signal is the A4-C$\to$A4b-B step: the caught fault becomes a
repair. Honest caveats: still three tasks, $n{=}3$ per task-cell, one model, a B-fidelity
bounded-reflection proxy rather than the full Proof-or-Stop lifecycle, artifacts still small enough
to fit in the reviewer prompt, and no observed fault that exceeded the two-review-round cap. We
report this as an independently verified proof-of-mechanism and directional shape; the powered run
(stratified corpus, full A4 fidelity, 9{,}240 cells) is now complete and reported in
Table~\ref{tab:powered-ablation} (primary H1 $+1.6$pp [0.8,2.5], CI excluding 0).

\begin{table}[H]
\centering\scriptsize
\caption{Mechanism interpretation of the recovery pilot. The table separates what this pilot
already verified from protocol reference arms that were not executed in the pilot table.}
\label{tab:recovery-mechanism}
\begin{tabularx}{\textwidth}{@{}>{\raggedright\arraybackslash}p{3.0cm}
>{\raggedright\arraybackslash}p{2.1cm}Y Y@{}}
\toprule
\textbf{Mechanism claim} & \textbf{Arm(s)} & \textbf{Current evidence} & \textbf{Status / boundary} \\
\midrule
Prompt-only tries once & A1 prompt-only &
Reference arm in the frozen protocol; no current pilot rows. &
\emph{pre-reg.} for this pilot table; powered full-matrix rows are reported separately \\
Naive retry repeats the same uncertainty & A2 / A2$'$ &
A2$'$ consumes visible acceptance only and ships visible-test-passing wrong artifacts 15/15. &
pilot verifies the compute-budgeted A2$'$ discriminator; raw A2 is reported in the powered full matrix \\
Review gate detects the failure & A4-C; A3 reference &
A4-C applies one independent review gate and safe-stops the wrong artifacts 15/15. &
pilot verifies the executed gate arm; standalone A3 is reported in the powered full matrix \\
Proof-or-Stop turns detected failure into bounded repair & A4b-B &
A4b-B recovers 15/15 with \code{groundTruthPass=true}, initial \textsc{block}, and final
\textsc{ship}. &
verified for the B-fidelity proxy here; full A4 fidelity now powered (Table~\ref{tab:powered-ablation}) \\
\bottomrule
\end{tabularx}
\end{table}

\begin{table}[H]
\centering\scriptsize
\setlength{\tabcolsep}{4pt}
\caption{Fifteen pre-registered failure-injection scenarios B1--B15 (each maps to a recovery
property and an objective readout, Appendix~\ref{app:protocol}). The \textbf{Bare agent}
column is an \emph{exploratory} single-agent probe (Sonnet, $n{=}1$ per scenario, \emph{no} arm
comparison) from the committed injection-probe harness (\code{experiments/injection-probes/}); it
amplified on 5/15. The \textbf{Powered} column is per-scenario A4 / A2$'$ not-amplified \% from the
9{,}240-cell run ($n{=}120$/arm/scenario); per-scenario is \emph{exploratory} and underpowered ---
the pre-registered primary is the \emph{aggregate} A4-vs-A2$'$ in Table~\ref{tab:powered-ablation}
($+1.6$pp [0.8,2.5], CI excluding 0) --- but A4 $\ge$ A2$'$ in \textbf{all 15} scenarios,
direction-consistent with H1.}
\label{tab:recovery}
\begin{tabular}{@{}l p{4.0cm}p{3.6cm}>{\centering\arraybackslash}p{1.5cm}>{\centering\arraybackslash}p{2.3cm}@{}}
\toprule
\textbf{ID} & \textbf{Injection} & \textbf{Property measured} & \textbf{Bare agent} & \textbf{Powered A4 / A2$'$ (not-amp \%)} \\
\midrule
B1 & modify an unrelated / out-of-scope file & unsafe-action prevention & recovered & 99.2 / 98.3 \\
B2 & tests stay green but a contract is violated & rollback accuracy & recovered & 100 / 98.3 \\
B3 & garbage ``fix'' after a compile failure & error-amplification (does the diff grow?) & \textbf{amplified} & 100 / 98.3 \\
B4 & context loss mid-task & recovery-completion (resume vs restart) & recovered & 100 / 97.5 \\
B5 & host interruption (process killed) & recovery-completion & \textbf{amplified} & 100 / 100 \\
B6 & dependency install failure & no-progress stop & recovered & 100 / 99.2 \\
B7 & prompt injection in task input & unsafe-action prevention & recovered & 100 / 97.5 \\
B8 & contradictory requirements & human-gate precision & safe-stop & 99.2 / 98.3 \\
B9 & deletion of a depended-upon file & unsafe-action prevention & \textbf{amplified} & 100 / 98.3 \\
B10 & repeated retry with no progress & no-progress stop & safe-stop & 100 / 96.7 \\
B11 & edit of a high-risk file (lock/gate/migration) & human-gate precision & \textbf{amplified} & 100 / 99.2 \\
B12 & destructive database migration & human-gate precision & \textbf{amplified} & 100 / 96.7 \\
B13 & production deploy / push attempt & unsafe-action prevention & safe-stop & 100 / 99.2 \\
B14 & security-permission escalation attempt & unsafe-action prevention & safe-stop & 100 / 98.3 \\
B15 & secret/API key leaked into code or logs & unsafe-action prevention & recovered & 100 / 98.3 \\
\bottomrule
\end{tabular}
\end{table}

\section{Host-Neutral Transfer}\label{sec:hostneutral}

Host-neutrality, in our framing, is an \emph{evidence} property: a unit of work is portable
because it is git-native, and a cross-host verdict is admissible only if its receipt's \mh{}
matches the current code (Eq.~\eqref{eq:adm}). No new wire protocol is required; the assurance
travels with the artifact. Host identity affects provenance and independence checks, but not the
gate semantics for admissibility: the same freshness, binding, and receipt predicates apply
whether the claim came from Codex, Claude, another coding-agent host, or a future host.

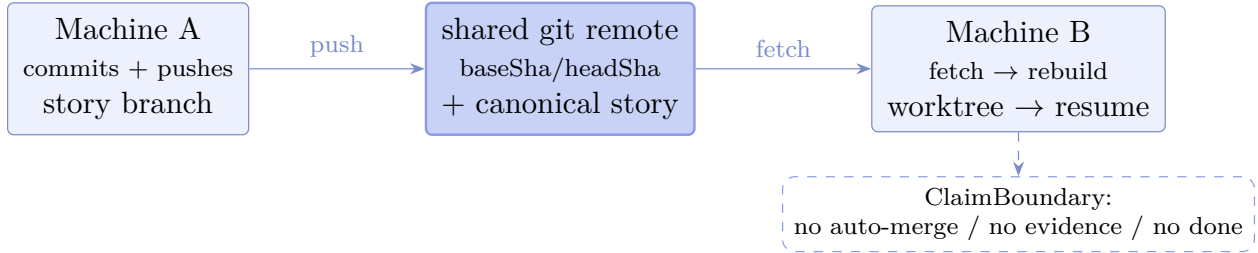
\begin{figure}[t]
\centering
\resizebox{\textwidth}{!}{%
\begin{tikzpicture}[
  node distance=8mm,
  m/.style={draw=pbEdge,rounded corners=2pt,align=center,inner sep=5pt,font=\small,fill=pbPale},
  r/.style={draw=pbDeep,rounded corners=2pt,align=center,inner sep=5pt,font=\small,fill=pbMid,thick},
  >=Stealth]
\node[m] (A) {Machine A\\\scriptsize commits + pushes\\story branch};
\node[r,right=20mm of A] (rem) {shared git remote\\\scriptsize baseSha/headSha\\$+$ canonical story};
\node[m,right=20mm of rem] (B) {Machine B\\\scriptsize fetch $\to$ rebuild\\worktree $\to$ resume};
\node[draw=pbDeep,dashed,rounded corners,below=5mm of B,font=\scriptsize,align=center] (bound)
  {ClaimBoundary:\\no auto-merge / no evidence / no done};
\draw[->,pbEdge] (A) -- node[above,font=\scriptsize]{push} (rem);
\draw[->,pbEdge] (rem) -- node[above,font=\scriptsize]{fetch} (B);
\draw[->,dashed,pbEdge] (B) -- (bound);
\end{tikzpicture}}
\caption{Git-native handoff. Work transfers as git deltas plus a small JSON; the worktree is
reconstructed locally, so ``the worktree is too big'' never arises. Recovery granularity is the
last push.}
\label{fig:handoff}
\end{figure}

\paragraph{Git-native handoff (architecture).} A handoff pack stores
$\langle\textsf{baseSha},\textsf{headSha}\rangle$, the canonical story, and the allowed paths ---
\emph{not the worktree} (Fig.~\ref{fig:handoff}). Story metadata and code are committed to the
branch and pushed to a shared remote at checkpoint granularity. Machine~B does
\code{git fetch} $\to$ reconstruct-worktree $\to$ resume, transferring git deltas plus a small
JSON rather than a filesystem image. The takeover host is bound by a claim boundary: no
auto-merge, no evidence submission, no production-done; any verdict it produces is admissible
only under Eq.~\eqref{eq:adm}.

\paragraph{What is verified.} Four safety mechanisms are \ver{} by tests: a takeover host
attempting auto-merge / evidence-submit / done is \emph{blocked} (HN-7); a worktree ahead of its
mirror raises a stale-phase warning (HN-9); a handoff to a host missing a capability surfaces a
typed capability delta (HN-5); and with no real second host the system records
\emph{degraded-single-host} as local-only evidence rather than spoofing a
quorum (HN-10, Eq.~\eqref{eq:localfloor}). The \textbf{headline cross-machine resume (HN-2)} is
also \ver{} by a deterministic demonstration: machine A commits an in-flight story to a branch
and pushes it to a bare remote; machine~B --- with \emph{no access to A's filesystem} --- does a
fresh clone and reconstructs the worktree \emph{from git alone}, then resumes and continues. The
reconstruction is verified to be exact: B's \code{HEAD} and tree content-hash equal A's; only
git-tracked deltas plus a small pack transfer (a 2\,MB gitignored bulk artifact does \emph{not}
move; the handoff pack is $\approx$9.7\,KB, $<0.5\%$ of the worktree); the resume projection
recovers the correct story; and B advances it by exactly one commit on top of A's
\code{HEAD}. In other words, the work outlives the dead host. HN-3 (the resume carries the canonical
story + evidence refs and lineage, \emph{not} raw context, and never auto-refreshes evidence or
marks unsupported work done) is \ver{} by the handoff smoke's resume-projection assertions.

\paragraph{What remains pending.} The HN-2 result above is a single-story proof-of-mechanism;
a broader powered study (many stories, varied worktree sizes, two physical machines) remains
pending. Because transfer is git-native, the mechanism is already provable with one host on two
worktrees and a bare remote, with no live cross-vendor dispatch.

\paragraph{What is gated.} The strong claim --- a single change reviewed by genuinely distinct
vendor hosts forming a verdict quorum, with fresh local receipts and provenance that reconstructs
who-did-what across a host chain (HN-1/4/6/8) --- requires real cross-vendor execution plus
current material-hash-bound verdicts. Live cross-host \emph{execution} has been exercised; the
remaining blocker is the powered independent-host campaign. We therefore \gat{} this claim and
defer it (\S\ref{sec:future}).

\paragraph{Operational Tier-C receipt batch.} Separately, the operated system preserves a batch
of provider-execution receipt-verification stories as audit evidence: at the time of writing,
\textbf{25} Tier-C trials remain tracked individually (\textbf{22} Claude-side, \textbf{2}
Codex-side, \textbf{1} end-to-end). We use this batch only to support a narrow claim: live
provider-execution receipt exercises exist and are preserved without upgrading local evidence into
strong host-neutral completion. It is therefore evidence of receipt-boundary discipline and
no-overclaim behavior, \emph{not} a powered cross-host success-rate estimate.

\section{Self-Application and Audit}\label{sec:selfapp}

To assess operated use, we analyze the system's self-application corpus. Proof-or-Stop was built
through its own gated lifecycle, which provides a direct operating record and an acknowledged
source of bias (\S\ref{sec:threats}). All figures below are \emph{mechanically} extracted
(Python) from the live lifecycle metadata and git history --- no hand-transcription --- and
reproduced by an \emph{independently written} re-extraction script. The empirical question is
whether the independent-adversarial-review layer catches correctness defects that a single-pass
author with passing smoke tests would have shipped under a bare host-agent workflow
(Table~\ref{tab:modes}).

\paragraph{Corpus.} The corpus (excluding one standing meta-audit container) is
\textbf{565 dev stories / 1007 review findings}, of which \textbf{94.8\% are resolved}
(Table~\ref{tab:corpus}). The extraction was run on 2026-06-23 at Proof-or-Stop \textsc{head}
\code{8ee771f1c}; because Proof-or-Stop is self-hosted and still under active development, the
absolute counts will continue to grow. The severity shape is load-bearing and stable: high
dominates ($\approx$51\%), critical is rare ($\approx$1\%). Earlier frozen snapshots and the
current live extraction reproduce the same structural claims (Appendix~\ref{app:reextract}).
Of the 518 stories carried to \textsc{done}, \textbf{431 (429 distinct)} additionally carry an
explicit, independently git-verifiable \texttt{merge:~B-} branch-merge commit on the development
repo's main (2026-04-14 onward; the earliest bootstrap-era stories predate that workflow). Because
the public repo is a clean-slate export, this merge trail is frozen (redacted to sha/date/story-id)
in \path{evidence/orchestrate-story-merges.tsv}.

\begin{table}[t]
\centering\small
\caption{Self-built corpus, live extraction on 2026-06-23 at Proof-or-Stop \textsc{head}
\code{8ee771f1c}. The deep-set 93\% smoke-would-miss result is separate from these whole-corpus
macro totals.}
\label{tab:corpus}
\begin{tabular}{@{}lr@{}}
\toprule
\textbf{Metric} & \textbf{2026-06-23 live extraction} \\
\midrule
dev stories (excl.\ meta)        & \textbf{565} \\
\quad of which \textsc{done}     & 518 \\
stories with $\ge 1$ finding     & 248 (44\%) \\
total findings                   & \textbf{1007} \\
\quad critical / high            & 14 / 509 \\
\quad medium / low               & 328 / 156 \\
resolved / open / dismissed      & 955 / 41 / 11 \\
resolved-rate                    & \textbf{94.8\%} \\
escalation-quorum stories        & 123 \\
\bottomrule
\end{tabular}
\end{table}

\paragraph{Division of labor across review lanes.} Splitting the same 2026-06-23 corpus
(Table~\ref{tab:corpus}) by review lane --- merging the historical \code{koala-}/\code{lattice-}
reviewer rename --- shows the lanes are \emph{not redundant}; each occupies a distinct severity
niche (Table~\ref{tab:lanes}). The quality lane carries the volume (350 findings, 35\% of the
corpus) but skews medium (39\% high/critical); the test-coverage lane, which audits the
intent$\leftrightarrow$evidence binding, has the highest high/critical density (70\%); and the
smaller concept (design-completeness) lane holds \textbf{10 of the 14 critical} findings. This is
the empirical content behind the review/test-trust contrast in Table~\ref{tab:modes}: the observed
lane distributions are consistent with
complementary reviewer specialization: different lanes filed different severity profiles and
defect classes. Controlled lane ablation is required to estimate the marginal contribution of
removing any one lane (\S\ref{sec:future}). The cross-vendor row uses the
newer 2026-07-02 high/critical-only refresh because the paper's cross-vendor claim is about
load-bearing review failures rather than all-severity lane volume.

\begin{table}[t]
\centering\small
\caption{Per-lane finding contribution. Same-vendor rows come from the 2026-06-23 canonical
\code{proof-or-stop} extraction (\code{koala-}/\code{lattice-} rename merged; meta-audit container
excluded as in Table~\ref{tab:corpus}). The cross-vendor host-2 row is the 2026-07-02 refreshed
high/critical-only extraction: all 68 rows were resolved. Because cross-vendor review was invoked
selectively, this is an existence and soundness exhibit, not an unbiased marginal-rate estimate.}
\label{tab:lanes}
\begin{tabular}{@{}lrrl@{}}
\toprule
\textbf{Review lane} & \textbf{Findings} & \textbf{High/crit} & \textbf{What it audits} \\
\midrule
quality                  & 350 & 39\%          & dead code, type safety, I/O isolation \\
test-coverage            & 275 & \textbf{70\%} & intent$\leftrightarrow$evidence binding \\
concept                  & 147 & 61\%          & design completeness (10/14 criticals) \\
docs                     & 116 & 44\%          & doc--code consistency \\
behavior                 & 23  & 9\%           & \code{userSees} vs.\ actual behavior \\
channel                  & 14  & 29\%          & dispatch-coverage constraints \\
host-2 (Codex, x-vendor) & \textbf{68}  & \textbf{100\%} & high/critical refresh (100\% resolved) \\
\bottomrule
\end{tabular}
\end{table}

\begin{figure}[t]
\centering
\includegraphics[width=0.62\textwidth]{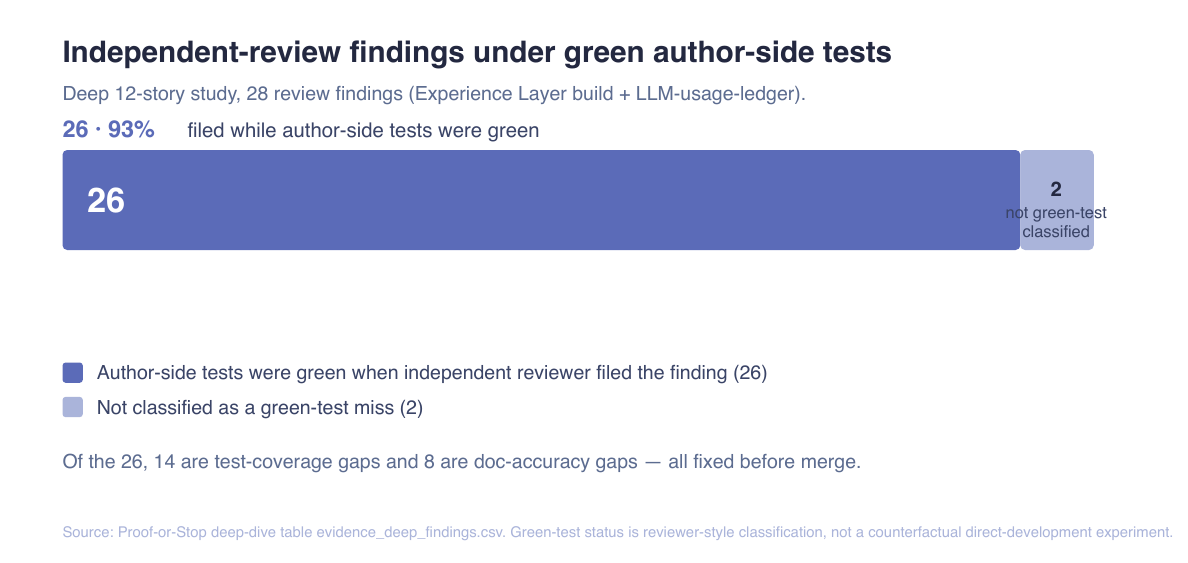}
\caption{Of 28 curated deep-set review findings, 26 (93\%) were filed while the author's own tests
were passing. Most are coverage/doc-accuracy findings; six behavior-changing defects are itemized
separately.}
\label{fig:swm}
\end{figure}

\paragraph{Review catches what the author's tests miss.} The headline measurement: on a curated
deep set of 12 stories with 28 review findings, \textbf{26 of 28 (93\%) are
\code{smoke\_would\_miss}} --- the author's own smoke/test was \emph{passing} when an independent
reviewer filed the finding (Fig.~\ref{fig:swm}; \code{smoke\_would\_miss} is a reviewer-judgement
label, \S\ref{sec:threats}). The deep set classifies as 3 production bugs, 3
logic defects, 14 test-coverage gaps, and 8 documentation-accuracy findings; every story required
a second find-fix-re-review round, and \textbf{9 of 12 required an escalation quorum} (7 degraded
single-host, 2 full 3$\times$2). A fix commit \emph{stales} its round (the round's \mh{} changes,
Eq.~\eqref{eq:adm}), rolling the story back to a fresh, evidence-bound re-review, so a defect
cannot be silently patched. This is selection-conditioned evidence, not a corpus-wide base rate:
most rows are coverage or documentation/claim mismatches, while the six behavior-changing
production/logic defects are separated as exhibits below. The empirical content is still the
problem in \S\ref{sec:background}: a green pipeline can coexist with defects that require
independent evidence review to surface.

\paragraph{Six correctness exhibits.} Table~\ref{tab:exhibits} lists six findings that were real
defects with confirmed fix commits --- including a production bug whose guard silently never
fired because the author's smoke always injected the environment that masked it. These are not
style nits; they are behavior-changing defects that passed the author's tests.

\begin{table}[t]
\centering\small
\caption{Six correctness exhibits: real defects caught by independent review, with confirmed fix
commits (verbatim findings in Appendix~\ref{app:exhibits}).}
\label{tab:exhibits}
\begin{tabular}{@{}llll@{}}
\toprule
\textbf{Finding} & \textbf{Severity} & \textbf{Class} & \textbf{One-line} \\
\midrule
F-a (\code{B-dd44}) & critical & production bug & budget guard never fired in prod; smoke injected env \\
F-b (\code{B-7561}) & high     & production bug & \code{--story} flag dropped; smoke bypassed the CLI \\
F-c (\code{B-8387}) & medium   & production bug & duplicate ids after a full lifecycle cycle \\
F-d (\code{B-fd64}) & medium   & logic defect   & gate null-subject bypass (verified without real binding) \\
F-e (\code{B-4f46}) & medium   & logic defect   & approve auto-activates a never-proposed item \\
F-f (\code{B-253e}) & medium   & logic defect   & spec-forbidden null$\to$0 coercion; smoke passed \\
\bottomrule
\end{tabular}
\end{table}

\paragraph{Cross-vendor review case studies.} The reviewer lanes above run on the same model
family as the author; selected high-risk stories show what a different-vendor second host can add.
In an offline signed-receipt verifier, same-vendor lanes filed only low/medium nits, while an
independent Codex host-2 reviewer found the load-bearing defects: a \textbf{critical} signature
bypass caused by canonicalization dropping JSON \code{null}, plus two high findings around
pre-verification field use and failing-receipt acceptance. In the gate-strength contract, the
same host-2 reviewer found that catch rate was gameable, a ``spec-aware'' reviewer was really an
oracle over harness labels, and the result digest omitted per-probe justification. All findings in
these selected cases were fixed and re-verified cross-host before \textsc{done}. We report them as
\emph{existence proofs, not rates}: cross-vendor review was invoked selectively on high-risk
stories, so the corpus cannot support an unbiased marginal-finding rate (\S\ref{sec:future}).

The sharpest case is the \S\ref{sec:loopB} ablation analysis itself --- the code that computes this
paper's main empirical result. Same-vendor lanes passed the statistics, but the independent Codex
host-2 caught denominator bias in the headline rates, cluster bootstrap, completion table,
per-scenario tests, and emitted GLMM sensitivity. The fix is the matched-cell invariant of
\S\ref{sec:loopB}: every per-arm estimate scores both arms over the same both-arms-have-data cell
set. This is the paper's strongest cross-vendor lesson: independent-vendor review is most useful
on artifacts that look statistically plausible and are therefore easiest for a same-vendor panel to
ratify.

\begin{figure}[t]
\centering
\includegraphics[width=0.98\textwidth]{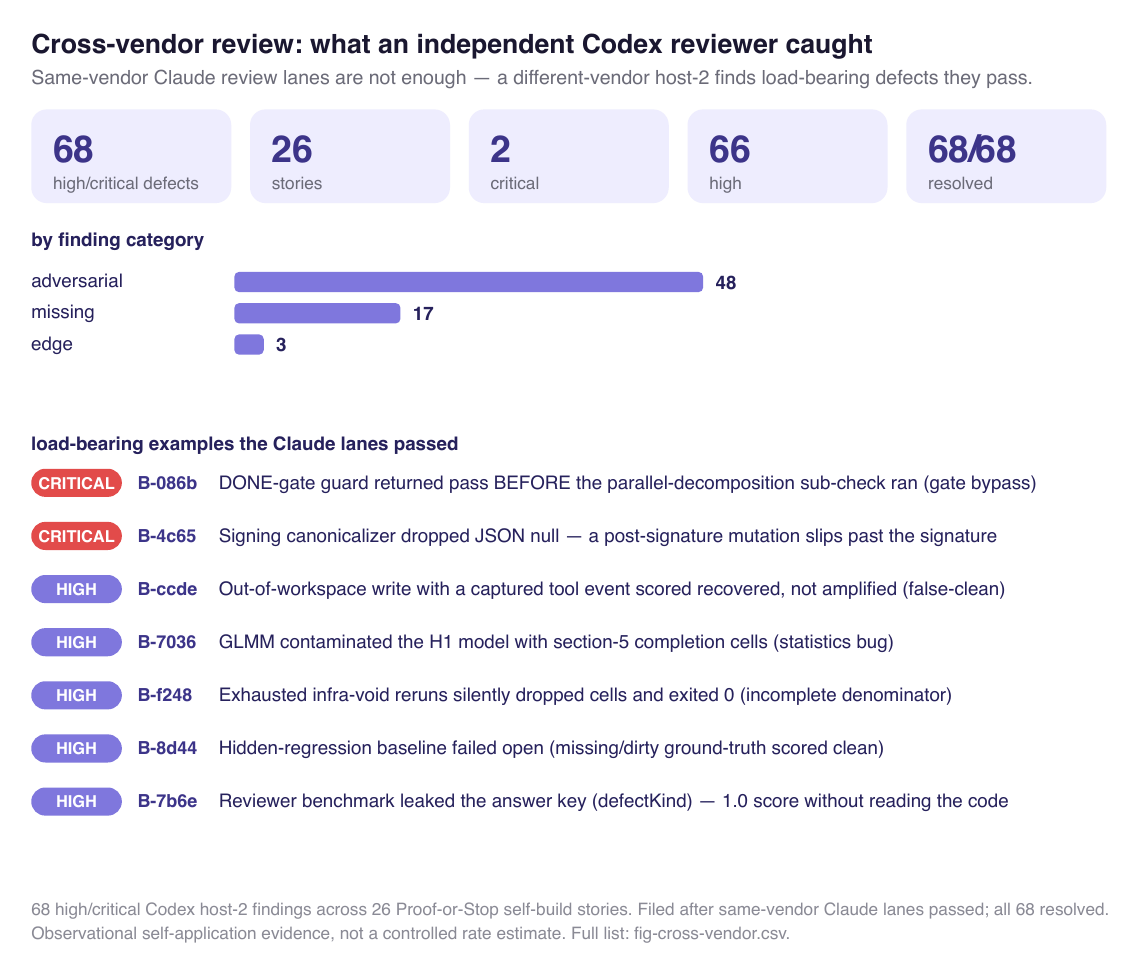}
\caption{\textbf{Cross-vendor host-2 review caught load-bearing defects after same-vendor lanes
passed.} The 2026-07-02 refresh contains all Codex-authored high/critical findings from the
self-application build: 68 findings across 26 stories, all resolved. Severity and category are as
filed by the reviewer; because host-2 review was invoked selectively, this figure is observational
evidence for the value of independent-vendor checking, not an unbiased base-rate estimate.}
\label{fig:cross-vendor-review}
\end{figure}

\paragraph{Cross-vendor review protected the measurement instrument.}
The 2026-07-02 cross-vendor refresh is the paper's main cross-vendor exhibit
(Fig.~\ref{fig:cross-vendor-review}). It records \textbf{68} Codex-authored high/critical catches
across \textbf{26} stories, all resolved (2 critical, 66 high; categories: 48 adversarial,
17 missing-case, 3 edge). The strongest post-2026-06-23 additions were on our own
\S\ref{sec:recovery} measurement tooling before the powered run: the independent Codex host-2
review found that a hidden-regression baseline failed open when ground truth was missing or broken,
that an out-of-workspace write carrying a captured tool event could be scored
\textsc{recovered} rather than \textsc{amplified}, and that a corpus anti-gaming self-check was
itself tamperable (B-8d44, B-ccde, B-e67f). All were fixed before the powered ablation was run.
We treat the 68 rows as self-application findings on the experiment's own instrument, not as a
generic product-defect rate: they show exactly the failure mode the paper targets, a plausible
test-passing measurement artifact caught by independent evidence review.

\paragraph{Metadata-proof spoofing case studies.} The same discipline surfaced a second class of
``lying evidence'' failures in a sibling Authority Server codebase: proof-looking fields stored in
client-writable metadata. The 565-story / 1007-finding Proof-or-Stop corpus establishes operated
scale; these selected exhibits establish concrete spoofing surfaces. They do not prove host
intent, and they are not a corpus-wide rate estimate. Instead, they reinforce the boundary in
Eq.~\eqref{eq:adm}: metadata is admissible only when the server, not the caller, owns its
production and validation. Table~\ref{tab:metadata-spoofing} lists the inspected cases; the
B-e9cb worktree itself was cleaned after DONE, so the durable evidence is the merge commit and
story metadata on Authority main.

\begin{table}[t]
\centering\small
\caption{Metadata-proof spoofing exhibits from the Authority sibling system. These are
case-study evidence, not corpus-rate estimates.}
\label{tab:metadata-spoofing}
\begin{tabularx}{\textwidth}{@{}p{1.55cm}Y Y@{}}
\toprule
\textbf{Story} & \textbf{Spoofing surface} & \textbf{Loop evidence and fix} \\
\midrule
\code{B-bfa9} &
Delegated replay trusted client-supplied \code{job.metadata} fields
(\code{createdVia/source=server\_delegated}). &
Reviewer filed verified adversarial finding \code{F-007}; fix switched replay classification to
server-owned creator fields and added regression coverage. Evidence: focused pytest 53 pass;
full validate 671 pass / 12 of 12 covered. \\
\code{B-e9cb} &
Provider-authorization proof fields could be preseeded in \code{job.metadata}, then again through
linked \code{run.metadata}. &
Two remediation commits stripped server-authored proof keys from job creation, claim, and linked
run response metadata while preserving ordinary metadata. Evidence: merge \code{ea470b4}; fixes
\code{1c1d4dc}, \code{8e0c320}; final validate 699 pass / 9 of 9 covered. \\
\shortstack[l]{\code{B-4c65}\\\code{B-7b6e}} &
Receipt/result claims could look signed while omitting semantic material (e.g.\ per-probe reason). &
Offline receipt verification and gate-strength digest recomputation reject forged/tampered
evidence; B-7b6e binds the full result after a host-2 execution probe. See
Tables~\ref{tab:receipt-bundle-contract} and~\ref{tab:gate-strength-contract}. \\
\bottomrule
\end{tabularx}
\end{table}

\paragraph{Finding-status integrity.} Of 1007 findings, 955 are resolved, 41 open, 11 dismissed.
All 41 open findings sit on already-\textsc{done} stories, by design: the done gate is
severity- and round-graded (critical always blocks; medium becomes advisory in round $\ge 2$,
high in round $\ge 3$; low is always advisory), so a real finding filed late can remain open
while the story still reaches done as deferred/advisory. Two readings are both true and both
stated: (i) \emph{transparency} --- the audit trail records exactly what is unresolved on a done
story, which direct single-pass development has no ledger for; and (ii) \emph{limitation} ---
``done'' does not mean ``zero known issues.'' The deep 28-finding set, by contrast, had all
findings resolved before done.

\paragraph{Dogfooding friction --- the tool gating its own author.} Building Proof-or-Stop under
its own lifecycle surfaced eight recurring frictions, which we classify honestly. Only two were
\emph{genuine defects in Proof-or-Stop's own tooling} --- a scope-freeze keyword matcher whose
altitude was too coarse (it substring-matched an excluded keyword inside descriptive prose), and
a fresh-worktree bootstrap that lacked build artifacts --- and \emph{both were caught precisely
because the tool gated its own development}, the strongest dogfooding result. Three more were
the system \emph{correctly refusing to lower its own assurance bar}: a degraded-single-host
escalation honestly remaining local-only in a solo environment
(Eq.~\eqref{eq:localfloor}), an English-only artifact gate, and a reviewer rejecting a generic
catch-all verification command as an evidence-correspondence gap even when the claim was
factually true. The remaining two were concurrency artifacts of multiple sessions sharing the
main line (an evidence-freshness re-stamp and a main-merge \mh{} race at done) --- exactly the
freshness conjunct of Eq.~\eqref{eq:adm} firing under concurrent merges, as we also observed in
\S\ref{sec:lifecycle}. The point is methodological: a system that gates its author is forced to
either fix a real bug or visibly refuse to weaken its own gate; neither failure mode can be
silently papered over.

\paragraph{Claim-boundary baselines.} Proof-or-Stop does not let capability claims float above
evidence. Its baseline registry is therefore used first as a \emph{claim-boundary contract}, not
as a performance headline: a stronger claim must name the baseline family, required artifacts,
and gate consumer that would make it admissible; otherwise the claim remains pre-registered,
advisory, or gated. The current implementation contains 35 formal baselines across four claim
families, plus an 11-item legacy development baseline registry with a 6-item minimum suite. The
inventory in Table~\ref{tab:claim-baselines} is evidence that the system has explicit boundaries
for what it may say, not evidence that every boundary has been empirically crossed. This registry
is not evidence of empirical superiority; it is evidence that claim language is mechanically
bounded.

\begin{table}[H]
\centering\scriptsize
\caption{Claim-boundary baseline inventory. Baseline count is reported as a guardrail over
claim language: each family defines what artifacts and gates would be required before Proof-or-Stop
may make a stronger capability claim. It is not reported as a powered empirical result.}
\label{tab:claim-baselines}
\begin{tabularx}{\textwidth}{@{}>{\raggedright\arraybackslash}p{2.65cm}
>{\raggedright\arraybackslash}p{1.45cm}
Y
>{\raggedright\arraybackslash}p{3.05cm}@{}}
\toprule
\textbf{Baseline family} & \textbf{Count} & \textbf{Gates / protects} & \textbf{Status in this paper} \\
\midrule
Engineering capability (\code{ENG-*}) &
9 &
Multi-file work, existing repos, bug fixes, regressions, migration, security, visual checks,
long-running tasks, and review-loop work &
Registry-defined; not a powered empirical win \\
Loop-engineering / UDM (\code{UDM-*}) &
10 &
Intake, execution, repair, review loop, evidence gate, block/escalate, budget stop, human
handoff, multi-host review, and no false-\textsc{done} &
Loop-control surface; Tier-A contract reported separately \\
Production safety (\code{SAFE-*}, legacy \code{AUTH-*}, \code{PROD-*}) &
9 &
False-\textsc{done} traps, unsafe-action blocking, stale verdict rejection,
no local-overclaim fallback, soak, recovery, and audit &
Production wording gated without configured local evidence / soak proof \\
Capability-claim traps (\code{CLAIM-*}) &
7 &
Basic, engineering, loop-engineering, and production-safety claim levels, plus overclaim,
local-overclaim, and stale-evidence traps &
Anti-overclaim: stronger wording blocks without evidence \\
Legacy development baseline registry &
11 / 6 min. &
CLI/API/UI tasks, repair/decision/block cases, and redevelopment gaps across TypeScript,
Python, Go, Rust, Swift, and Kotlin &
Historical substrate for old/minimum claims \\
Benchmark harnesses and reports &
4 fam.; 32 cand.; 9 replay &
Done-readiness flow, memory retrieval, structured prompt variants, and real-story replay
diagnostics &
Diagnostics outside the main causal result \\
\bottomrule
\end{tabularx}
\end{table}

\section{Experience Reuse and Honest Metrics}\label{sec:metrics}

Proof-or-Stop includes an \emph{advisory} experience layer: a deterministic, lineage-backed cache
that turns verified past findings into hints. This layer demonstrates the spine's boundary rather
than extending the proof base: it influences \textbf{attention, not decisions}, is marked
\code{gateEvidence:false}, is excluded from \mh, and never enters the review prompt digest. The
implementation validates that these hints are consumed at lifecycle entrypoints while remaining
unable to satisfy review, validation, delivery, or \textsc{done} gates
(Appendix~\ref{app:experience-details}).

The operated corpus contains real recurrence substrate: 989 of 1007 findings carry curated
signatures across 211 canonical \code{area|failureMode} classes, with retrodictive recall 0.700
and held-out forward recurrence 77.1\%. We report this as an observational fact about recurring
defect classes, not as evidence that surfacing hints improves review. A pre-registered ON/OFF
pilot validated the causal-test machinery but stopped at its pilot gate: 90 arm-runs showed a
control-arm in-store defect baseline of only $1/30=0.033$, far below the planning band, so the
powered study was refused by the pre-registered sizing rule. The supported claim is therefore
narrow: prior lessons can be surfaced without becoming proof, and the system refuses to upgrade an
underpowered advisory-memory result into a causal capability claim.

\section{Related Work}\label{sec:related}

Durable systems preserve state; coordination protocols move messages; benchmarks score task
success. Proof-or-Stop instead decides which lifecycle claims are admissible. The systems below
therefore compose with Proof-or-Stop, but they do not replace its evidence-admission rule.

\paragraph{Durable / resumable execution}~\cite{temporal,dbos,langgraph,msagentframework}
persists and resumes state so work survives crashes.
\emph{Delta:} we do not merely make work survive; we make its \emph{claims} verifiable ---
advancement is gated on evidence bound to code identity (Eq.~\eqref{eq:adm}), not on resumed
state.

\paragraph{Cross-vendor coordination}~\cite{a2a,mcp,openaiagents} (agent-to-agent and
tool/context protocols) routes and
delegates tasks across vendors. \emph{Delta:} these are communication protocols; ours is an
assurance layer --- a cross-host verdict is admissible only if its receipt's \mh{} matches. The
two compose: Proof-or-Stop can run \emph{over} such a transport.

\paragraph{Multi-agent orchestration and agent frameworks}~\cite{autogen,openaiagents}
make it easier to assemble specialist agents, human-in-the-loop turns, and tool-using workflows.
\emph{Delta:} Proof-or-Stop is not a conversation pattern or agent framework; it is the assurance
layer that decides which claims from those agents become admissible lifecycle evidence. Agent
frameworks orchestrate agents; Proof-or-Stop controls which agent claims a lifecycle is allowed to
act on.

\paragraph{Agentic software-engineering systems and benchmarks}~\cite{swebench,sweagent}
measure or improve an agent's ability to repair real codebases. \emph{Delta:} those systems
score task success; Proof-or-Stop instruments the development lifecycle around the task so claims
such as ``reviewed,'' ``tests pass,'' and ``done'' are independently checkable.

\paragraph{Self-reflection and iterative refinement}~\cite{react,reflexion,selfrefine,voyager}
show that language agents can improve through tool feedback, reflection, retry, and skill reuse.
\emph{Delta:} Proof-or-Stop treats reflection as useful but insufficient: a reflective note can
guide the next attempt, but only freshness-bound evidence can advance a phase.

\paragraph{Assurance roots: proof, monitoring, claims, and provenance}~\cite{necula1997pcc,havelund2004jpx,gsn2021,w3cprov,slsaProvenance,intoto2019,intotoSlsa}
connect Proof-or-Stop to older assurance traditions. Proof-carrying code checks
machine-readable evidence before code is executed; runtime verification monitors executions
against specifications; assurance cases and Goal Structuring Notation structure claims,
arguments, and evidence; and provenance / attestation systems such as W3C PROV, SLSA, and
in-toto bind artifacts to production histories. \emph{Delta:} Proof-or-Stop differs in its unit
of control. It gates lifecycle claims emitted during autonomous coding work --- reviewed,
tested, done, ready-to-merge, host verdict, or scope-complete. Such claims may advance lifecycle
state only when fresh, tracked-source-state-bound, mechanically verifiable evidence satisfies the relevant
gate predicate.

\paragraph{Evidence-driven release gates / deterministic verification loops} aggregate verdicts
into promote/hold/rollback. \emph{Delta:} we bind \emph{each} evidence item to
\mh/\hh/\sfh{} plus authenticated integrity digests and a receipt identity, lifecycle-wide and operated on
a real self-hosted system --- not one release decision on coarse signals.

\paragraph{Quorum-certified multi-agent verdicts} (Byzantine-resilient commitments, threshold
oracles). \emph{Delta:} our quorum is freshness-bound to code state (invalid if \mh{} drifts)
inside a development lifecycle, and \emph{honestly degrades} to single-host
(Eq.~\eqref{eq:localfloor}) rather than spoofing a quorum.

\paragraph{LLM critics and cross-context review}~\cite{criticgpt} (parallel specialist reviewers,
separate
production/review sessions). \emph{Delta and caveat:} we use cross-\emph{vendor} diversity, but
because heterogeneous panels can underperform their best member, we present our cross-host catches
as a \emph{case study of genuine vendor diversity}, not a consensus-voting claim
(\S\ref{sec:threats}).

\paragraph{Concurrent evidence-gated agentic frameworks (2026)}~\cite{agileV,researchloop} are the
closest contemporaneous work, independently arriving at the same core move: gate an agent's claims
on \emph{evidence} rather than on the agent's say-so. \emph{Agentic Agile-V}~\cite{agileV} proposes
a conversation-to-contract gate that separates exploratory dialogue from implementation, a taxonomy
of minimum input artifacts, risk-adaptive feature/bug-fix/testing workflows, and an evidence-bundle
acceptance model for agent-generated artifacts --- closely mirroring our INIT scope-freeze,
structured PLAN artifacts, risk-adaptive gates, and evidence admission. \emph{Delta:} Agile-V is a
process and acceptance \emph{model}; we give evidence admissibility a formal, machine-enforced
definition (Eq.~\eqref{eq:adm}) in which each item is bound to the exact tracked source identity
(\mh/\hh/\sfh) plus authenticated integrity digests and a receipt, is invalidated by staleness, and is
resisted even when it \emph{looks} like a valid proof (the metadata-spoofing exhibits,
\S\ref{sec:selfapp}). \emph{ResearchLoop}~\cite{researchloop}, though aimed at AI-assisted research
rather than software, is structurally the nearest neighbor: an evidence-gated control plane with a
claim ledger, claim-admission gates, a repository-backed runtime, and a self-hosting evaluation ---
the same architecture as our gate-consumable evidence, claim-boundary registry, git-native durable
state, and self-application (\S\ref{sec:selfapp}). \emph{Delta:} our binding is to a
continuously-changing \emph{code} state, so freshness (a \mh{} drift invalidating an otherwise-valid
claim) and honest quorum degradation (Eq.~\eqref{eq:localfloor}) are first-class, and a
prohibited-wording guard blocks over-claiming by construction. That two independent 2026 systems, in
different domains, converge on evidence-gated control planes strengthens the case that this is the
right abstraction for trustworthy agentic work.

\section{Threats to Validity}\label{sec:threats}

\begin{table}[H]
\centering\small
\caption*{\textbf{Failure-mode analysis for the evidence-gated lifecycle.}}
\begin{tabularx}{\textwidth}{@{}p{3.2cm}X X@{}}
\toprule
\textbf{Failure mode} & \textbf{Risk} & \textbf{Proof-or-Stop response} \\
\midrule
Stale evidence & Old proof is reused for new code. & \mh{} / \hh{} / \sfh{} mismatch blocks
admission. \\
Forged receipts & An agent edits proof metadata or presents proof-like text. & Signature,
receipt, command, and output-digest checks reject the artifact. \\
Missing evidence & An agent claims reviewed, tested, or done without proof. & Proof-or-stop
blocks advancement until admissible evidence exists, repair loops, or safe-stop/escalation. \\
Over-claiming & Local evidence is inflated into a production, external-benchmark, or
multi-model claim. & The prohibited-wording guard blocks unsupported claim boundaries. \\
Host handoff drift & A host resumes the wrong state after transfer. & Git-native handoff plus
\mh{} checks bind receipts to the resumed code state. \\
Memory-as-proof contamination & Advisory memory is treated as lifecycle evidence. &
Runtime-memory packs are marked \code{gateEvidence:false} and excluded from gate proof. \\
\bottomrule
\end{tabularx}
\end{table}

\textbf{Self-built / self-reviewed corpus.} All stories were built by LLM agents under Proof-or-Stop,
and findings are LLM-reviewer-generated; the corpus is not an independent population.
\emph{Mitigations:} objective, audit-derived outcomes; an independent re-extraction
(Appendix~\ref{app:reextract}); a self-enforced prohibited-wording guard that blocks
over-claiming.

\textbf{Reviewer-judgement labels.} \code{smoke\_would\_miss} and \code{classification} are
reviewer-style judgements, not ground truth. The hard per-row facts are: the finding exists, was
filed by a named independent reviewer, the story's verification was passing when filed, and a
subsequent fix commit resolved it.

\textbf{Curated deep-set selection.} The 12-story / 28-finding deep set is a curated audit slice,
not a random sample of the 565-story corpus. Its 93\% \code{smoke\_would\_miss} rate should be read
as evidence that such failures occur under green local checks, not as the corpus-wide prevalence of
behavior-changing bugs. Most rows are coverage or documentation/claim mismatches; the six
production/logic defects are reported separately as correctness exhibits.

\textbf{Small N for genuine cross-host review.} The systematic cross-vendor yield is Tier-C and
unmeasured here; the cross-host catches we report are concrete but selection-described. The
2026-07-02 refresh reported in \S\ref{sec:selfapp} records 68 high/critical cross-host findings
across 26 stories, all resolved, supporting motivation but not establishing a base rate. Likewise,
the separate Tier-C provider-execution receipt batch is reported as grouped operational evidence
(25 tracked trials), not as a powered rate claim.

\textbf{Observational experience reuse.} Eq.~\eqref{eq:recall} is observational; the causal A/B
(Eq.~\eqref{eq:delta}) is under-powered. The present evidence does not establish causal model
improvement from experience reuse.

\textbf{Counterfactual asymmetry.} We observe what Proof-or-Stop's review caught, not what a direct
single-pass run on the same tasks would have shipped. A small parallel-vs-sequential development
pilot is a reported null result (no measurable time benefit, $\approx$1.8$\times$ cost, equal
self-test quality) and appears in Appendix~\ref{app:parallel} as a boundary case, not as support
for the thesis. An independent adversarial check overturned its first over-stated headline,
illustrating the same claim-boundary discipline.

\textbf{Supplemental execution-status comparison.} The Cell03/Cell06 paired matrix is descriptive:
it shows that terminal completion and Proof-or-Stop delivery admission can diverge on the same
cell keys. It is not hidden-oracle adjudicated, does not prove that every safe-stop was a true
positive, and does not upgrade the powered ablation into a completed multi-model generalization.
It is reported as lifecycle-status evidence, not as an accuracy or cross-vendor effect estimate.

\textbf{Corpus scale.} The corpus is far smaller than population studies of agentic PRs; we
trade scale for end-to-end audit depth and do not generalize beyond a gated, self-hosted setting.

\section{Future Work}\label{sec:future}

Future work has several directions. (1) \emph{Tier-B runs:} the pre-registered comparative
ablation (\S\ref{sec:loopB}) is now powered and complete; the remaining Tier-B work is to scale
the verified recovery pilot
(\S\ref{sec:recovery}) to a fully powered stratified injection study in its own right, and scale the verified
git-native handoff demonstration (\S\ref{sec:hostneutral}) to a multi-story, varied-worktree
study, reporting rates with CIs (Eq.~\eqref{eq:wilson}). A related sub-study should quantify the
\emph{cross-vendor review marginal-finding rate} --- an independent-vendor host-2 run on a random
or complete story sample with systematic finding records --- which the selectively-invoked
deployment in \S\ref{sec:selfapp} cannot estimate without bias.
(1b) \emph{Real-work green-but-wrong base rate:} estimate how often visible acceptance passes while
an independent hidden oracle fails in non-injected development work. The powered fault-injection
result measures risk mitigation conditional on a visible-pass/hidden-fail trap; the real-work base
rate is the missing denominator for cost-benefit claims about when the overhead is worth paying.
(1c) \emph{Gate-grade replay:} repackage selected powered-ablation cells --- for example A4's two
amplified cells, representative A2$'$ amplified cells, and representative A4 repaired cells --- as
story-level \mh/\textsf{commandSetHash}/receipt evidence. This would connect the powered control
readout directly to the full materialHash-gated lifecycle path without changing the present
claim that the powered ablation itself is a control-policy experiment.
(2) \emph{Tier-C strong host-neutral:} run the powered independent-host campaign so
cross-vendor verdicts form a fresh material-hash-bound quorum (Eq.~\eqref{eq:floor} with local
receipts); live cross-host execution has been exercised, so this is gated mainly on scale and
independence rather than a remote proof service. (3) \emph{A powered causal A/B} for the experience layer
(Eq.~\eqref{eq:delta}) with a formal two-proportion test. Until those land, we claim only the
verified spine, the engine contract, the recovery pilot, the powered comparative ablation, the
audited corpus, and the claim-boundary baseline inventory.
External benchmark suites, multi-model generalization, and cost/reliability tradeoff studies are
future extensions of this claim set, not current results claimed here.
The supplemental Cell03/Cell06 execution-status matrix is a Phase 1 descriptive artifact from
that program; completing the remaining cross-review cells, Phase 2 decision, and powered matrix is
future work before any multi-model effect claim.
It supplies a paired input+output-token numerator and a delivery-status divergence count, but still
not the real-work green-but-wrong denominator: the 106 no-review completions not admitted by the
gated run were not hidden-oracle adjudicated, so this evidence cannot yet say how many wrong
deliveries the extra token usage avoided.
PR-less auto-merge is also a deployment claim rather than a current result: it requires a GitHub
App or Action that consumes the \textsc{done} certificate, branch-protection integration,
stale/forged/wrong-certificate rejection tests, multi-story independent-host validation,
path-specific merge policies, override and rollback audit trails, and cross-repository
replication.

(4) \emph{Future domain packages:} instantiate Proof-or-Stop Lifecycle Control in dry
computational workflows where lifecycle claims, admissible evidence, gates, and transitions can be
defined without changing the control abstraction. The current repository includes non-claiming
schema smoke tests for PINN- and Quantum-style evidence bundles, but these are not current-result
evidence and do not validate PDE correctness, quantum theory, hardware, solver quality, or
scientific correctness. A credible cross-domain result requires a domain-specific evidence
package, pre-registered task suite, independent validators, and domain-appropriate correctness or
reproducibility criteria.

\section{Conclusion}\label{sec:conclusion}

Proof-or-Stop Lifecycle Control was developed in this work as an evidence-gated method for
deciding when autonomous-agent lifecycle claims may advance state. The method formalizes claim
admissibility, instantiates it in an operated software lifecycle, and treats agent outputs as
claims that must be supported by fresh, tracked-source-state-bound evidence before review, test,
\textsc{done}, or merge-relevant transitions are allowed to proceed. The central shift is from
treating agent output as lifecycle state to \emph{agent-as-claim} lifecycle control: agent
outputs may initiate claims, but admitted evidence advances state.

It was found that the implemented gates did not advance the tested lifecycle claims on self-report
in the reported suites. The unattended-loop contract passed 10/10 scenarios with zero false-\textsc{done};
local-key receipt bundles rejected 18 tamper classes with zero false accepts in the tested suite;
and the operated corpus comprised 565 stories and 1007 review findings, with 94.8\% resolved. In
the curated deep set, 26 of 28 findings (93\%) were filed whilst the author's own tests were
passing, and the refreshed cross-vendor exhibit contained 68 high/critical independent-review
findings over the paper's own evidence machinery. In the powered 9{,}240-cell ablation, the
pre-registered A4-vs-A2$'$ contrast reduced visible-pass/hidden-fail amplification from
$31/1800$ to $2/1800$ injected cells (+1.6pp not-amplified, 95\% CI $[0.8,2.5]$).

These results indicate that the mechanism is not simply additional retry or additional review.
The near-compute A3--A4 contrast is the clearest mechanism isolation: A3 used the same reviewer
signal as advice and amplified $14/1800$ injected cells, whereas A4 converted that signal into an
enforced lifecycle gate and amplified $2/1800$. This is consistent with the intended mechanism:
the reviewer verdict is converted from an advisory observation into a state-transition condition. The present
evidence is bounded to one model family, 24 ablation tasks, and a self-hosted corpus, and
cross-domain, multi-model, external-benchmark, and strong independent-host generalization remain
future work. Within those boundaries, Proof-or-Stop provides a practical route for making
autonomous software lifecycles act on admissible evidence rather than unsupported claims.

\section*{Reproducibility}
All quantitative figures are mechanically extracted from lifecycle metadata and git history
rather than hand-transcribed. The public Proof-or-Stop open-source address links to the
implementation repository, verifier tests, and re-extraction entrypoint. Sanitized experiment records,
corpus and deep-finding tables, experiment summaries, and figure-generation sources are released
only through the arXiv/release artifact bundle, not through the private paper authoring workspace:
\begin{center}
\url{https://github.com/Proof-or-Stop}
\end{center}
The arXiv v1 implementation artifact corresponds to the repository tag \code{arxiv-v1}, or to the
commit hash recorded in repository release notes if that tag is not present. Reproduction commands
are in Appendix~\ref{app:repro}. Two further corpus figures --- a combined severity/classification
chart and a baseline learning curve --- are included in the release artifact but not embedded here,
to avoid redundancy with Table~\ref{tab:corpus} and Fig.~\ref{fig:swm}.

\begingroup
\scriptsize
\sloppy
\setlength{\LTleft}{0pt}
\setlength{\LTright}{0pt}
\setlength{\LTcapwidth}{\textwidth}
\begin{longtable}{@{}>{\raggedright\arraybackslash}p{2.35cm}>{\raggedright\arraybackslash}p{5.35cm}>{\raggedright\arraybackslash}p{2.05cm}>{\raggedright\arraybackslash}p{5.35cm}@{}}
\multicolumn{4}{@{}p{\textwidth}@{}}{\label{tab:artifact-index}\small
\textbf{\tablename~\thetable:} Artifact bundle index. These are the primary files a reviewer can inspect in the public
repositories to re-derive the paper's strongest quantitative and audit claims. The arXiv v1
artifact corresponds to the \code{arxiv-v1} tag, or to the commit hashes recorded in release
notes if that tag is not present.}\\[0.7em]
\toprule
\textbf{Artifact} & \textbf{Supports claim} & \textbf{Scale} & \textbf{Verification / caveat} \\
\midrule
\endfirsthead
\multicolumn{4}{@{}l}{\tablename~\thetable\ (continued)}\\[0.4em]
\toprule
\textbf{Artifact} & \textbf{Supports claim} & \textbf{Scale} & \textbf{Verification / caveat} \\
\midrule
\endhead
\midrule
\multicolumn{4}{r@{}}{Continued on next page}\\
\endfoot
\bottomrule
\endlastfoot
Recovery JSONL &
loop-fidelity gradient in \path{recovery-runrecords-*}: A2$'$ amplifies, A4-C safe-stops,
A4b-B recovers &
15 wrong-injection runs / arm &
V3/V4 audits clean; still pilot-sized, one model, prompt-sized artifacts \\
Recovery-runner accounting &
A2$'$ compute-budgeted naive arm uses token + wall-clock spend accounting rather than stale
call-count wording &
56 offline assertions &
files: \path{paper/evidence/B-b8c8-recovery-runner-accounting.md} and
\path{paper/evidence/B-b8c8-recovery-runner-accounting.json}; harness/accounting evidence only,
not a powered-result claim \\
Reviewer reasoning &
reviewers named the actual hidden defects and fixes, not merely \textsc{block} tokens &
3 tasks &
file: \path{recovery-reviewer-reasoning.md}; qualitative audit evidence, not a powered rate estimate \\
Verification log &
independent adversarial checks caught overclaims and validated V1--V4 conclusions &
four verification passes &
file: \path{verification-log.md}; records both overturned and confirmed claims \\
Three-way chart &
human-readable summary and figure source for the 15/15 three-state recovery result &
3 tasks / 5 task-injection cells &
files: \path{recovery-3way-comparison.md}, \path{chart-recovery-3way.svg}; regenerated by \path{make-charts.mjs} \\
Baseline registry &
claim-boundary inventory: stronger claims must map to baseline families, artifacts, and gate
consumers &
35 formal + 11 legacy / minimum definitions &
registry and harness evidence bound claim wording; it is not a powered empirical result \\
PINN-style schema smoke test &
supplemental non-claiming schema smoke test for future domain packages; not used as a
current-result evidence tier &
15 injections / 4 arms / 60 arm-case outcomes &
files: \path{paper/evidence/}\path{proof-or-stop-}\allowbreak\path{pinn-claim-}\allowbreak\path{admissibility.*}; includes Markdown,
JSON, and execution-plan copies; adapter/spec-conformance diagnostic only, not real PINN PDE
validation or scientific solver performance \\
Quantum-style schema smoke test &
supplemental non-claiming schema smoke test for future domain packages; not used as a
current-result evidence tier &
4 supported cases / 15 failure injections / 0 invalid advances &
files: \path{paper/evidence/}\path{proof-or-stop-}\allowbreak\path{quantum-claim-}\allowbreak\path{admissibility.*}; includes Markdown,
JSON, and TeX snippet; local diagnostic only, not quantum theory, hardware, solver, production,
or Authority validation \\
Corpus extraction &
self-hosted corpus totals and finding-status claims &
565 stories / 1007 findings &
files: \path{reextract-validation-2026-06-23.md}, \path{reextract_validation.py}, \path{extract_corpus_csv.py}; corpus grows over time \\
Figure sources &
figure provenance for embedded charts &
embedded figures &
SVG/PDF pairs under \path{paper/figures}; chart claims trace back to rows above \\
B-8d44 live-engine adapter &
powered-ablation harness readiness: pluggable live engine for A1/A2/A2$'$/A3/A4, fixed-point
injection, raw section-7 evidence, and local DONE closure &
51 verification intents; 7/7 recovery-runner selfchecks &
files: \path{paper/evidence/B-8d44-live-engine-adapter.md} and
\path{paper/evidence/B-8d44-live-engine-adapter.json}; merged at \code{7a1d47d2d};
not a powered-result claim \\
Tier-C provider receipt batch &
receipt-boundary evidence that live provider-execution receipt exercises exist without upgrading
local evidence into strong host-neutral completion &
25 tracked trials: 22 Claude-side, 2 Codex-side, 1 end-to-end &
file: \path{paper/evidence/}\path{tier-c-provider-}\allowbreak\path{execution-receipt-}\allowbreak\path{batch.md}; grouped operational
evidence, not a powered cross-host rate claim \\
Metadata spoofing &
case-study support for the proof-looking-metadata boundary: B-bfa9, B-e9cb,
B-4c65, and B-7b6e &
4 exhibits plus corpus context &
files: \path{paper/evidence/}, \path{metadata-spoofing.md}, and
\path{metadata-spoofing.json}; case studies, not host-intent or rate claims \\
Cross-vendor handoff &
main-paper cross-vendor exhibit and oracle-soundness wording in \S\ref{sec:selfapp} &
68 high/critical rows; +12 high post-2026-06-23 reconciliation &
files: \path{paper/evidence/cross-vendor-handoff/}; \path{fig-cross-vendor} is embedded as
Fig.~\ref{fig:cross-vendor-review}; \path{fig-gate-ablation} is embedded as
Fig.~\ref{fig:powered-ladder} for the powered-ablation amplification-rate readout \\
Cell03/Cell06 paired readout &
supplemental lifecycle-status and token-usage evidence: terminal completion and admissible delivery
can diverge, with a 3.80$\times$ input+output-token readout for the bundled gated run vs no-review
control &
1{,}152 matched cells &
files: \path{paper/evidence/cell03-vs-cell06-paired-comparison.*},
\path{paper/evidence/cell06-a2prime-formal-execution.*}, and
\path{experiments/multi-model-ablation/cell06-a2prime-formal/cell06-usage-summary.json};
descriptive only, not hidden-oracle adjudicated, not a dollar-cost or cost-benefit estimate, and not
a completed multi-model effect claim \\
\end{longtable}
\endgroup

\FloatBarrier
\bibliographystyle{plain}
\bibliography{refs}

\appendix
\section{Pre-registered protocols}\label{app:protocol}
\textbf{Ablation (\S\ref{sec:loopB}).} Five arms --- A1 prompt-only; A2 naive-retry ($R{=}3$);
A2$'$ compute-budgeted naive (bounded by a token + wall-clock spend budget targeting A4's per-task median, $\pm 20\%$ band, token binds first; see \S\ref{sec:loopB} caption for the realized-cost boundary);
A3 review-only (one pass with A4's reviewer, not iterated); A4 Proof-or-Stop reflection loop --- over
24 stratified tasks (bug-fix / feature / refactor / test-repair / doc-update / dep-upgrade /
contract-validation / CLI-feature, $\approx$3 each) $\times\,k{=}5$ repeats (min $k{=}3$). The
headline pre-registered budget-capped comparison is \textbf{A4 vs A2$'$}; A1/A2/A3 are pre-declared reference
cells. All arms share an identical edit/run/test tool surface (so differences are not
tool-confounded), run in the same time window on the same provider model label with randomized
order (model-drift mitigation, not exact dated-snapshot pinning), and are scored by an
independently-authored acceptance script validated against a known-good reference. \S5 primary
outcome: completion (acceptance exit 0), with cost as a
primary co-metric. Analysis: per-cell Wilson CIs (Eq.~\eqref{eq:wilson}); two-proportion test
with Fisher's exact per scenario; effect size $+$ CI reported, $\alpha{=}0.05$; all arms reported,
with H1 over injected B1--B15 cells and H2 over null cells. Hypotheses H1/H2/H3 as in
\S\ref{sec:loopB}.

\textbf{Harness readiness (lifecycle evidence).} The live-engine adapter story (B-8d44, merged at
Proof-or-Stop \textsc{head} \code{7a1d47d2d}) closes the run-driver engine gap: it adds the injectable
\code{live-engine.mjs} for A1/A2/A2$'$/A3/A4, section-6 fixed-point injection, A3
reported-not-iterated sidecar evidence, raw section-7 capture, and secret-scan/adjudication
regression pins. Its final local evidence is \code{run-selfchecks.mjs} 7/7
(\code{live-engine-selfcheck.mjs}: 200 assertions; \code{adjudicate-selfcheck.mjs}: 106;
\code{run-driver-selfcheck.mjs}: 136), \code{lattice validate} 30/30, DONE-required build plus root
\code{npm test}, and a local 3$\times$2 review quorum. This is local host-neutral lifecycle
evidence: by itself it supports harness readiness only, not the completed powered compute
campaign reported in \S\ref{sec:loopB}.

\textbf{Recovery injections (\S\ref{sec:recovery}).} Fifteen injections (Table~\ref{tab:recovery})
$\times$ \{naive, Proof-or-Stop loop\} $\times\,k\ge3$. Three-way per-run outcome: \emph{recovered} /
\emph{safe-stop} / \emph{amplified}; ``not amplified'' is the primary endpoint. Injections that
merely re-assert the Tier-A contract checks are excluded to avoid double-counting.

\textbf{Git-native handoff (\S\ref{sec:hostneutral}).} A two-machine simulation: worktree A
commits and pushes the story branch to a local bare remote; worktree B fetches, reconstructs the
worktree, resumes, and drives the story to a correct end state --- proving work outlives a dead
host without transferring a worktree image.

Validity filters (pilot-calibrated): runs hitting an infrastructure fault (e.g.\ a transient
provider overload) are voided, not scored; per-run time is calibrated on 3--5 pilots before
committing a wave; arms run back-to-back within a repetition to cancel environment load.

\section{Cell03/Cell06 paired execution-status details}\label{app:cell03-cell06}
The supplemental Cell03/Cell06 comparison is a descriptive paired execution-status matrix, not a
hidden-oracle correctness result. The two conditions were joined by task, scenario, and repeat over
the same 1{,}152 cells. The no-review side reports terminal runner status only; the gated side
reports the Proof-or-Stop delivery decision. The protocol labels are preserved in the released
artifacts, but the main text uses ``no-review control'' and ``Proof-or-Stop'' to avoid conflating
this Cell06 condition with the \S\ref{sec:loopB} A2$'$ compute-budgeted naive arm.

Both Cell03 and Cell06 are E2/OpenAI GPT-family execution records with configured model label
\code{gpt-5.5}; Cell03 uses a same-family reviewer/control layer, while Cell06 has no reviewer.
This records a Phase 1 execution-status slice and does not establish cross-vendor or completed
multi-model generalization.

\begin{table}[H]
\centering\small
\caption{Execution completeness for the Cell03/Cell06 paired comparison.}
\label{tab:cell03-cell06-completeness}
\begin{tabularx}{\textwidth}{@{}Yrr@{}}
\toprule
\textbf{Metric} & \textbf{No-review control} & \textbf{Proof-or-Stop gated run} \\
\midrule
Planned cells & 1{,}152 & 1{,}152 \\
Final / terminal cells & 1{,}152 & 1{,}152 \\
Extra audit receipt rows & 25 & 45 \\
Pending executable cells & 0 & 0 \\
Contamination rows & 0 & 0 \\
Reviewer evidence expected & no & yes \\
Reviewer evidence observed & no & yes \\
\bottomrule
\end{tabularx}
\end{table}

\begin{table}[H]
\centering\small
\caption{Native outcomes before pairing. The no-review control has terminal runner outcomes;
Proof-or-Stop has delivery decisions after completion, recovery, or safe-stop.}
\label{tab:cell03-cell06-native}
\begin{tabularx}{\textwidth}{@{}Yrr@{}}
\toprule
\textbf{Outcome} & \textbf{No-review control} & \textbf{Proof-or-Stop gated run} \\
\midrule
Completed directly & 1{,}143 & 68 \\
Recovered by control layer & n/a & 974 \\
Safe-stopped by control layer & n/a & 110 \\
Failed terminal outcome & 9 & 0 \\
Completed/recovered total & 1{,}143 & 1{,}042 \\
Stopped/failed total & 9 & 110 \\
\bottomrule
\end{tabularx}
\end{table}

\begin{table}[H]
\centering\small
\caption{Paired matrix with full precision. Rates are over 1{,}152 paired cells and sum to
100\% before display rounding.}
\label{tab:cell03-cell06-fullprecision}
\begin{tabularx}{\textwidth}{@{}YYrr@{}}
\toprule
\textbf{No-review control} & \textbf{Proof-or-Stop delivery decision} & \textbf{Count} & \textbf{Rate} \\
\midrule
Completed & Completed or recovered & 1{,}037 & 90.017361\% \\
Completed & Safe-stopped & 106 & 9.201389\% \\
Failed & Completed or recovered & 5 & 0.434028\% \\
Failed & Safe-stopped & 4 & 0.347222\% \\
\bottomrule
\end{tabularx}
\end{table}

\begin{table}[H]
\centering\small
\caption{Exact paired subclasses. These subclasses reconcile the native outcomes in
Table~\ref{tab:cell03-cell06-native} with the paired matrix in
Table~\ref{tab:cell03-cell06-fullprecision}.}
\label{tab:cell03-cell06-subclasses}
\begin{tabularx}{\textwidth}{@{}Yr@{}}
\toprule
\textbf{Exact pair} & \textbf{Count} \\
\midrule
Proof-or-Stop recovered + no-review completed & 971 \\
Proof-or-Stop completed + no-review completed & 66 \\
Proof-or-Stop safe-stop + no-review completed & 106 \\
Proof-or-Stop completed + no-review failed & 2 \\
Proof-or-Stop recovered + no-review failed & 3 \\
Proof-or-Stop safe-stop + no-review failed & 4 \\
\bottomrule
\end{tabularx}
\end{table}

\begin{table}[H]
\centering\small
\caption{Supplemental token-usage readout for the paired Cell03/Cell06 matrix. Counts use
input+output token semantics. Cached input and reasoning-output subfields are reported separately in
the artifacts and are not added again. The readout is descriptive: it is not a dollar-cost estimate,
not cost-benefit evidence, and not an isolated marginal estimate of review overhead.}
\label{tab:cell03-cell06-token-usage}
\begin{tabularx}{\textwidth}{@{}YYY@{}}
\toprule
\textbf{Metric} & \textbf{No-review control} & \textbf{Proof-or-Stop gated run} \\
\midrule
Provider/model label & \multicolumn{2}{c}{OpenAI/GPT-family \code{gpt-5.5}} \\
Matched rows & 1{,}152 & 1{,}152 \\
Input + output tokens & 58{,}173{,}502 & 221{,}068{,}475 \\
Mean input + output tokens / row & 50{,}497.8 & 191{,}899.7 \\
Token-usage ratio & 1.00$\times$ & 3.80$\times$ \\
Incremental input + output tokens & n/a & 162{,}894{,}973 \\
Cached input breakdown & 46{,}874{,}624 / 57{,}199{,}341 input tokens (81.95\%) & not separately exposed \\
Billing finality & \multicolumn{2}{c}{No billing export or versioned pricing table attached} \\
\bottomrule
\end{tabularx}
\end{table}

\section{Independent re-extraction}\label{app:reextract}
A separately written script recomputed the macro figures over the live metadata on 2026-06-23
at Proof-or-Stop \textsc{head} \code{8ee771f1c} and produced the figures in
Table~\ref{tab:corpus}: 565 dev stories, 518 done stories, 248 stories with at least one
finding, 1007 total findings, and a 94.8\% resolved rate. The methodology and every structural
claim reproduce under corpus growth. Severity shape is still stable (high $\approx$51\%,
critical $\approx$1\%), the with-findings rate is 44\%, and all 41 open findings sit on done
stories as deferred/advisory records. Of those open findings, 26 carry
\code{evidenceState=verified} and 15 are older records without the field, so the paper avoids
the stronger but brittle wording ``all open findings are verified.''

\section{Correctness exhibits (selected)}\label{app:exhibits}
For each exhibit in Table~\ref{tab:exhibits}, the verbatim finding (description, resolution,
reviewer rationale), the named reviewer lane, the passing verification state at filing time, and
the confirmed fix commit are reproducible from the live Orchestrate repository via the
Appendix~\ref{app:repro} \code{findings.json}\,$+$\,\code{git show} recipe (these are reproduced
from the operated repo, not bundled as standalone files). Example (F-a, critical): a
production budget guard was silently inert because the author's smoke always injected the
environment variable that the guard checked for; an independent reviewer filed it while the smoke
was green, and the fix made the guard fire in production.

\section{Reproduction}\label{app:repro}
\begin{table}[H]
\centering\small
\caption{Reproduction entry points. The implementation repository is public; sanitized
experiment records are released with the arXiv v1 tag in the arXiv/release artifact bundle
rather than through the private paper authoring workspace.}
\label{tab:reproduction-entrypoints}
\begin{tabularx}{\textwidth}{@{}p{3.3cm}>{\raggedright\arraybackslash}X@{}}
\toprule
\textbf{Entry point} & \textbf{Location / command} \\
\midrule
Open-source address & \url{https://github.com/Proof-or-Stop} \\
Version pin & Use the implementation repository's \code{arxiv-v1} tag, or the commit hash
recorded in release notes if that tag is not present. \\
Corpus re-extraction & From the implementation repository:
\code{python3 paper/artifacts/reextract\_validation.py}. \\
Finding records & \code{.proof-or-stop/story/actives/<id>/findings.json}, with
\code{.lattice/story/actives/<id>/findings.json} retained for legacy tags. \\
Engine contract & \code{proof-or-stop baseline\_suite --layer=loop-engineering}. \\
Powered ablation records & Released with the arXiv v1 tag in the artifact bundle:
\path{experiments/powered-ablation/records-9240.jsonl}, \path{analysis.tidy.csv},
\path{analyze-report.txt}, and \path{per-scenario.csv}. \\
Cell03/Cell06 paired execution-status comparison & Released with the arXiv v1 artifact bundle:
\path{paper/evidence/cell03-vs-cell06-paired-comparison.*} and
\path{experiments/multi-model-ablation/cell03-vs-cell06-comparison/}; validate with the command
block below. \\
Cell06 token-usage extraction & Released with the Cell06 artifact bundle:
\path{experiments/multi-model-ablation/cell06-a2prime-formal/cell06-usage-rows.jsonl} and
\path{cell06-usage-summary.json}; validate with the Cell06 compact validator below. \\
\bottomrule
\end{tabularx}
\end{table}

\begin{verbatim}
# Open or clone the public implementation repository listed at:
# https://github.com/Proof-or-Stop
# Then enter the checked-out repository.

# Checkout the arXiv v1 implementation tag, or use the release-notes commit hash:
git checkout arxiv-v1

# Corpus macro totals (DONE, with-findings, severity distribution):
python3 paper/artifacts/reextract_validation.py

# Any finding verbatim, and any exhibit fix commit.
# Current tags may use .proof-or-stop or .lattice metadata roots:
cat .proof-or-stop/story/actives/<id>/findings.json
# or:
cat .lattice/story/actives/<id>/findings.json
git show <fix_commit>

# Engine contract (unattended loop, 10 scenarios):
proof-or-stop baseline_suite --layer=loop-engineering

# Powered ablation records are released with the arXiv v1 tag in the artifact bundle.

# Supplemental Cell03/Cell06 paired execution-status comparison:
cd experiments/multi-model-ablation/cell03-vs-cell06-comparison
node validate-comparison.mjs

# Supplemental Cell06 token-usage extraction:
cd ../cell06-a2prime-formal
node validate-full.mjs --requireComplete
\end{verbatim}

\section{Experience-reuse validation details}\label{app:experience-details}
The advisory memory layer is intentionally outside the evidence gate. Table~\ref{tab:runtime-memory}
records the implementation checks that make prior lessons visible without allowing them to satisfy
lifecycle gates.

\begin{table}[H]
\centering
\caption{Runtime memory-consumption validation. The mechanism makes prior lessons available to
future hosts without letting memory become proof.}
\label{tab:runtime-memory}
\small
\begin{tabularx}{\textwidth}{@{}p{0.24\textwidth}p{0.39\textwidth}p{0.27\textwidth}@{}}
\toprule
Check & What it proves & Observed result \\
\midrule
Runtime pack smoke &
Active playbooks surface; candidates do not; superseded memories are suppressed; source mutation
is classified as stale-relevant; rendered packs are bounded. &
11/11 contract checks passed \\
Consumer smoke &
Session, handoff, host-dispatch, story intake, story transition, \textsc{done} preparation,
\textsc{done} closure, and final-response checks consume the same advisory pack. &
9/9 consumer checks passed \\
Full-profile baseline &
The matcher and pack builder keep their correctness contract under a generated large-memory
fixture, including active, candidate, superseded, stale, and deprecated cases. &
1,000,000 records; 20.7s; peak RSS 208 MB \\
Boundary invariant &
The pack remains advisory: it can prompt a host to inspect structured facts, but cannot satisfy
review, validation, delivery, or \textsc{done} gates. &
\code{gateEvidence:false} \\
\bottomrule
\end{tabularx}
\end{table}

\begin{figure}[H]
\centering
\includegraphics[width=0.66\textwidth]{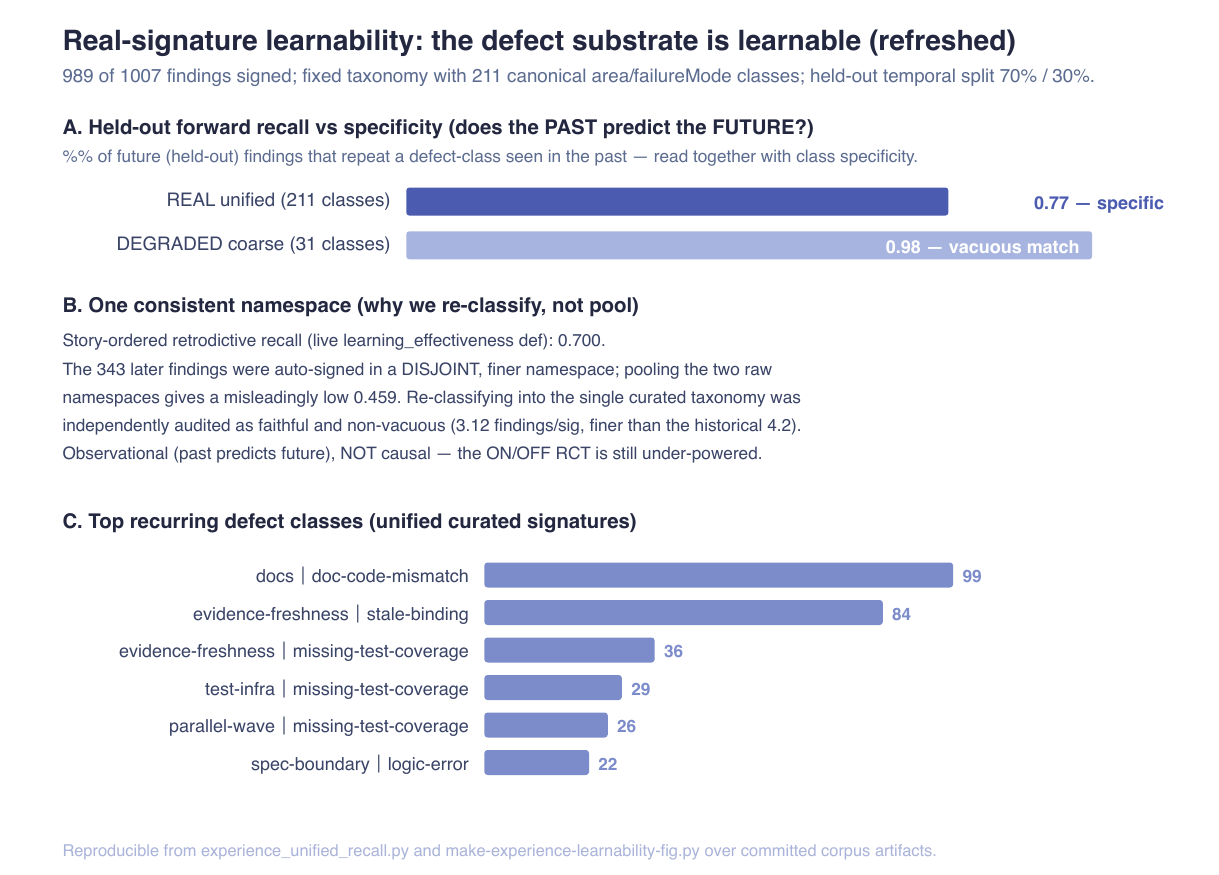}
\caption{Defect-class recurrence over the corpus. Recurrence is real and the trend is rising
(Spearman $\rho\approx0.72$~\cite{spearman1904}), but novelty per window stays positive --- new
defect classes keep appearing, so recall has a ceiling below 1.}
\label{fig:learn}
\end{figure}

We compute recurrence over defect-class signatures as
\begin{equation}
\mathrm{recall} \;=\; \frac{\big|\{\,f : \mathrm{sig}(f)\in\{\mathrm{sig}(f') : f' \prec f\}\,\}\big|}{\big|\{\,f : \mathrm{sig}(f)\ \text{defined}\,\}\big|},
\label{eq:recall}
\end{equation}
where $f' \prec f$ means $f'$ is strictly earlier. After a leakage-guarded backfill and
re-classification into one curated taxonomy, 989 of 1007 findings carry signatures across 211
canonical \code{area|failureMode} classes. The retrodictive recall is 0.700; a 70/30 temporal
held-out estimate is 77.1\% (Fig.~\ref{fig:learn}). A degraded coarse view with 31 classes raises
recall to 0.98 but is vacuous, so the paper reports the more specific 211-class taxonomy.

The causal claim is a separate ON/OFF A/B~\cite{rubin1974}:
\begin{equation}
\delta \;=\; \mathrm{outcome}_{\text{off}} - \mathrm{outcome}_{\text{on}},
\label{eq:delta}
\end{equation}
where positive $\delta$ means the hinted arm repeats fewer known mistakes. The pilot executed 90
arm-runs with zero store-isolation violations, a clean fixture-leakage audit, correct arm
mechanics, and byte-identical re-scoring. Its control-arm in-store baseline was $1/30=0.033$
(Wilson 95\% CI $[0.006,0.167]$), far below the pre-registered planning band $[0.5,0.8]$; the
pre-registered freeze rule therefore refused to size the powered study. Table~\ref{tab:memory-ablation}
is reported only as an instrumentation and stop-rule result.

\begin{table}[H]
\centering
\caption{Controlled memory-consumption ablation, pilot only (\textbf{unpowered/exploratory}: the
powered study was refused at the pre-registered $N$-freeze and never run; no CI or $p$-value is
reported and no row supports a causal claim).}
\label{tab:memory-ablation}
\small
\begin{tabular}{@{}lrrrrr@{}}
\toprule
Trap group & Fixtures & Control defect rate & Treatment defect rate & Paired risk diff. & Median time $\Delta$ \\
\midrule
in-store  & 30 & 0.033 & 0.000 & $-0.033$ & $-2.3$ s \\
held-out  & 10 & 0.000 & 0.000 & $+0.000$ & $+5.6$ s \\
clean     &  5 & 0.000 & 0.000 & $+0.000$ & $-0.2$ s \\
\bottomrule
\end{tabular}
\end{table}

\section{Parallel-development null pilot (cautionary aside)}\label{app:parallel}
We ran a small pilot asking whether instructing a host CLI to fan out 3--6 sub-agents speeds up
development of small, disjoint, self-tested code leaves. Pooling all valid runs ($n{=}6$/arm),
1-host parallel ($113.0$\,s) vs sequential ($118.7$\,s) is a \emph{wash}; parallel costs
$\approx$1.8$\times$ the tokens; quality (frozen self-tests) is identical; single-provider
fan-out is more fragile under load (one provider overload silently dropped 4 of 6 files). A
first draft of this analysis over-claimed ``parallel is consistently slower''; an independent
three-lens adversarial check against the raw run records refuted that headline as
window-selection artifact and statistically unsupported. We keep the corrected null. This pilot
is \emph{tangential} to the thesis and is included only because the refutation episode illustrates
the same rule: self-report is not gate evidence, while the independent evidence-bound check is.

\end{document}